\documentclass[letterpaper, 10 pt, journal, twoside]{IEEEtran}
\IEEEoverridecommandlockouts
\usepackage{amsmath,amsfonts, amssymb}
\usepackage{algpseudocode}
\usepackage[linesnumbered, ruled, vlined]{algorithm2e}
\usepackage{array}
\usepackage[caption=false,font=normalsize,labelfont=sf,textfont=sf]{subfig}
\usepackage{textcomp}
\usepackage{stfloats}
\usepackage{url}
\usepackage{verbatim}
\usepackage{graphicx}
\usepackage[hidelinks]{hyperref}
\usepackage{cite}
\usepackage{enumitem}
\usepackage{siunitx}
\usepackage{newtxtext}
\usepackage{booktabs}
\usepackage{listings}
\usepackage{xcolor}
\usepackage{multirow}

\usepackage{pifont} 
\newcommand{\orcid}[1]
{\href{https://orcid.org/#1}{\includegraphics[width=0.6em]{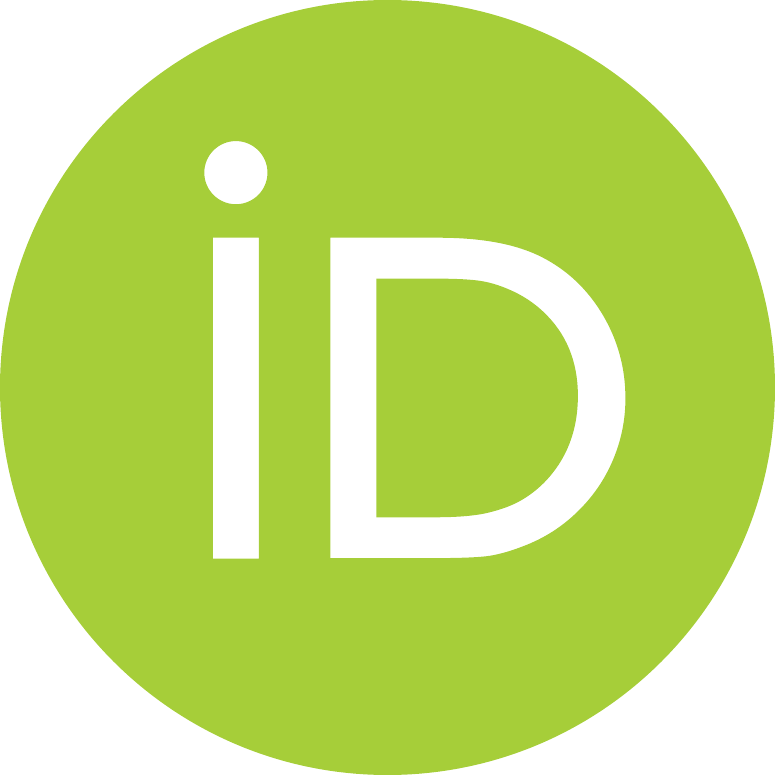}}}

\usepackage{glossaries}

\newif\ifpublished
\newcommand{\published}[1]{%
  \ifnum\pdfstrcmp{#1}{True}=0\relax
    \publishedtrue
  \else
    \publishedfalse
  \fi
}
\published{True} 

\begin{document}

\title{\textsc{CaLiSym}: Learning Symplectic Dynamics of Real-World Systems through Structured Canonical Lifts}
\ifpublished
\author{
    Aristotelis Papatheodorou${}^{*,1}$\orcid{0000-0003-0290-7071}\quad
    Pranav Vaidhyanathan${}^{*,1}$\orcid{0000-0003-4758-4210}\quad\\
    Natalia Ares$^{1}$\orcid{0000-0003-2588-6322}\quad
    Ioannis Havoutis$^{1}$\orcid{0000-0002-4371-4623}\quad
    Gerard J. Milburn$^{2}$\orcid{0000-0002-5404-9681}
\thanks{$^{1}$Aristotelis Papatheodorou, Pranav Vaidhyanathan, Natalia Ares and Ioannis Havoutis are with the Department of Engineering Science, University of Oxford, Oxford, UK. A.P. is supported by Oxford's Clarendon Fund and the JPMorgan Chase AI PhD Fellowship. P.V. is supported by the United States Army Research Office under Award No. W911NF-21-S-0009-2. N.A. acknowledges support from the European Research Council (Grant~agreement~948932) and the Royal Society (URF-R1-191150).}
\thanks{$^{2}$School of Mathematics and Physics, University of Sussex, Brighton, BN1 9RH, UK.  National Centre for Quantum Computing, Rutherford Appleton Laboratory,
Harwell Campus, Didcot, Oxfordshire, OX11 0QX UK}
\thanks{\vspace{-2mm}\\${}^*$Equal Contribution \texttt{\{aristotelis, pranav\}@robots.ox.ac.uk}}
\thanks{$^{\dagger}$Open-source implementation will be released upon acceptance.}
}
\else
\author{Author Names Omitted for Anonymous Review.
\thanks{$^{\dagger}$Open-source implementation will be released upon acceptance.}
}
\fi

\ifpublished
\markboth
{Papatheodorou \& Vaidhyanathan \MakeLowercase{\textit{et al.}}: \textsc{CaLiSym}}
{}
\else
\markboth
{Submitted to IEEE Transactions on Robotics.}
{Anonymous \MakeLowercase{\textit{et al.}}: \textsc{CaLiSym}: Learning Symplectic Dynamics of Real-World Systems through Structured Canonical Lifts}
{}
\fi


\maketitle

\begin{abstract}
Physics-informed learning promises data-efficient and stable dynamics prediction, yet its strongest geometric guarantees have largely remained confined to closed conservative systems. This excludes robotic systems of interest, where actuation, dissipation, and constraints exchange energy and momentum with the environment. We introduce \textsc{CaLiSym}, a lightweight framework that extends symplectic learning to such systems by changing where the geometric prior is imposed. Rather than enforcing symplecticity on the measured state, \textsc{CaLiSym} embeds the state and its ports into a lifted phase space, where the dynamics evolve through a symplectic map. The lift is explicit and algebraic, requiring neither recurrent latent states, transformer decoders, implicit optimization, nor inference-time numerical integration. We instantiate the framework with \textsc{SympNet} predictors and introduce \textsc{GRB-SympNet}, a B-spline variant combining approximation with exact symplectic structure. Experiments on a controlled dissipative double pendulum, a real-world quadrotor, and a contact-constrained real-world quadruped demonstrate the lowest out-of-distribution autoregressive rollout error across systems, improving by up to $\mathbf{69.5\%}$ while using fewer parameters and up to $\mathbf{85\times}$ fewer floating-point operations per step than sequence-model baselines. The lifted dynamics preserve the symplectic form to numerical precision, extending symplectic learning beyond conservative mechanics toward real-world robotics.
\end{abstract}
\vspace{0.1em}
\vspace{-0.3em}
\vspace{-3mm}
\section{Introduction}
\IEEEPARstart{D}{ynamics} has been studied extensively, producing a wide range of methods for explaining, predicting, and controlling the physical systems around us. The arrival of computation extended that reach to systems of scale and complexity that no closed-form analysis could touch. Among other fields, this shift has been highly consequential for robotics, where accurate dynamics models now underpin nearly every aspect of estimation, prediction and control. However, accuracy almost always comes with a cost.

Classical model-based control~\cite{connell2012robot, murray2017mathematical} rests on first-principles descriptions of the system, such as rigid-body mechanics, contact models, and actuator dynamics~\cite{featherstone2014rigid}. Yet these descriptions face two limits at once, one computational and one more fundamental. The computational limit surfaces in methods such as model-predictive control~\cite{mastalli2022agile} and state estimation~\cite{barfoot2024state}, which must repeatedly query the underlying model. High-fidelity models are often too expensive to evaluate at control rates, leading practitioners toward elaborate hand-engineered models~\cite{dai14centroidal, papatheodorou2024momentum} or coarse simplifications that widen the simulation-to-reality gap~\cite{wieber06holonomy}. The second limitation is more fundamental. Clean analytical models capture the dominant inertial, gravitational, and Coriolis effects, but high-performance robots operate precisely where neglected effects matter~\cite{selfridge1985training}. Aerial robots face thrust limits and aerodynamic drag~\cite{nguyen21aerial}, while legged robots face friction, compliance, impacts, and intermittent contact~\cite{tedrake23underactuated}. These are exactly the non-conservative effects that are hardest to derive from first principles and, as we will argue, the ones that most strongly violate the assumptions behind today's most reliable learned models~\cite{ai2025review, tsuji2026contact, ha2025learning,liu2024physics}.
\begin{figure}[t]
  \centering
  \includegraphics[width=0.8\columnwidth]{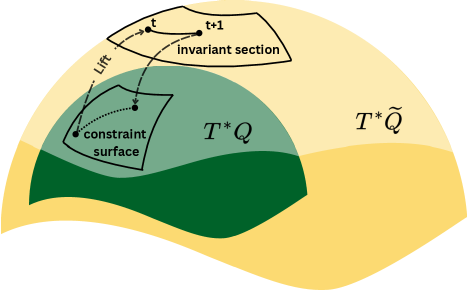}
  \caption{Overview of \textsc{CaLiSym}${}^{\dagger}$. The measured state lives in the physical phase space $T^*\mathcal{Q}$, the cotangent bundle of the configuration manifold $\mathcal{Q}$. Because the system exchanges energy and impulses through actuation, dissipation, and contact, the physical dynamics on $T^*\mathcal{Q}$ need not be closed or symplectic. \textsc{CaLiSym} therefore embeds the physical state $\mathbf{x}_t=(\mathbf{q}_t,\mathbf{p}_t)$ and its current ports $\mathbf{e}_t$ into a gauge-fixed data section $\mathcal{S}_{\mathbf{e}_t}$ of the lifted phase space $T^*\widetilde{\mathcal{Q}}$. An exactly symplectic map $\Phi_\theta^{\Delta t}$ advances the lifted state from $\mathbf{Z}_t$ to $\bar{\mathbf{Z}}_{t+1}$, after which the prediction is projected back to the physical state $\hat{\mathbf{x}}_{t+1}=\Pi_{\mathcal{X}}(\bar{\boldsymbol{Z}}_{t+1})$. For autoregressive rollout, the prediction is re-embedded with the next port value through $\mathbf{Z}_{t+1}=\sigma_{\mathbf{e}_{t+1}}(\hat{\mathbf{x}}_{t+1})$. Thus, symplecticity is enforced in $T^*\widetilde{\mathcal{Q}}$, while the projected physical dynamics on $T^*\mathcal{Q}$ may remain forced, dissipative, or contact-constrained.}
  \label{fig:lift}
  \vspace{-6mm}
\end{figure}

With the emergence of data-driven methods, predicting dynamics has shifted from hand-derived equations toward learned world models~\cite{ha18worldmodels}, i.e., neural predictors trained to roll out the future state of a system directly from data. In model-based reinforcement learning~\cite{sutton2018reinforcement, moerland2022mbrl}, such models are no longer passive simulators but are queried in closed loop, generating the imagined trajectories on which planning and policy optimization depend~\cite{hafner2020dreamer}. Their promise is speed without the brittleness of hand-tuned simplifications, with a single forward pass replacing an expensive solve. Their peril is that a learned map can be arbitrary. Small single-step errors compound over a rollout, and without any constraint grounding the prediction to the underlying physics, long-horizon trajectories drift and violate conservation laws the real system obeys. Their performance degradation is only amplified if we consider distribution shifts and conditions out of their nominal training scenarios~\cite{karoly2020deep}.

This is the gap physics-informed learning sets out to close~\cite{karniadakis2021physics, li23pinnoperators}. Rather than being left to learn an unconstrained transition map, the network is restricted to a model class built from known physical structure, so that its dynamics respect the physics by construction rather than only on average. In robotics, structure-preserving neural models have shown that such geometric priors improve data efficiency, extrapolation, and long-horizon stability~\cite{cranmer2020lagrangian, greydanus2019hamiltonian, roth2026stable, jin2020sympnets}. Yet the strongest of these guarantees comes with a catch. Exact symplecticity holds only for closed conservative systems, while the systems we want to control are forced, dissipative, and contact-constrained. As shown in Fig.~\ref{fig:lift}, we resolve this by imposing symplecticity on the right space, embedding the physical state and its ports into a structured lifted space $T^{*}{\tilde{\mathcal Q}}$ where the learned map is exactly symplectic, while the projected physical dynamics it induces reside in $T^{*}\mathcal Q$ and may be constrained, forced or dissipative.

The practical stakes of this construction extend beyond prediction accuracy. Learned dynamics models are increasingly queried in closed loop, serving as the transition model inside model-predictive control and imagination-based policy optimization, where compounding rollout error is the binding constraint~\cite{hafner2020dreamer, hansen2024tdmpc2}. This is where geometric fidelity earns its keep. A predictor that remains geometrically anchored over long horizons, while staying Markovian and requiring only a single forward pass, makes learned models usable inside such planners rather than merely alongside them.

The interface matches how modern robot controllers are actually built. The model predicts from the current state and ports alone, exactly the information a state-feedback controller or estimator holds at runtime, and the ports coincide with the joint torques and contact forces that predictive controllers already treat as decision variables~\cite{mastalli2022agile}. A receding-horizon solver can query the lifted predictor directly, with no observation history to maintain and no context window to replay at every iteration. Robustness under distribution shift matters for the same reason. A deployed robot routinely meets terrains, payloads, and contact modes absent from training, and the out-of-distribution gains reported here target precisely this regime.

Our framework offers a second advantage beyond stability. Because the learned map is exactly symplectic and the ports enter as explicit canonical variables, the model exposes the energy exchange between robot and environment as a first-class quantity rather than an implicit byproduct of the fit. This opens a path toward passivity-based and certificate-based control synthesis~\cite{vanderschaft2017l2gain, kushwaha2026review}, where energy bookkeeping is the object on which guarantees are built. Training stays equally practical, since teacher forcing avoids backpropagation through time and its attendant memory and gradient pathologies, keeping the model cheap to retrain as new operating data arrive. The predictor achieves all of this at parameter counts and inference costs compatible with control rates on embedded hardware.

\vspace{-3mm}
\subsection{Contributions}
This work introduces \textsc{CaLiSym}, a framework for learning Canonical Lifted Symplectic representations of real-world robot dynamics. The central idea is to enforce exact symplectic structure in a structured lifted phase space while allowing the induced physical dynamics to remain actuated, dissipative, and contact constrained. This bridges the gap between the theoretical guarantees of symplectic learning and the non-conservative dynamics encountered in practical robotic systems. Particularly, we make the following contributions:

\begin{itemize}

\item \textbf{Symplectic learning beyond conservative systems.}
We introduce \textsc{CaLiSym}, a framework that extends symplectic learning to controlled, dissipative, and constrained robotic systems (e.g. contacts).

\item \textbf{Structured lift and gauge fixed rollout.}
We introduce a structured lift that augments the physical state with variables governing energy exchange with the environment and constraints such as contacts. A gauge-fixed lift, projection, and re-embedding procedure enables autoregressive prediction under time varying controls and intermittent contacts. Due to the enforced symplectic priors the autoregressive error accumulation remains minimal, while the overall Markovian rollouts remain stable.

\item \textbf{\textsc{GRB-SympNet}.}
We introduce \textsc{GRB-SympNet}, a generalized ridge B-spline symplectic architecture that combines local spline approximation with exactly symplectic flow layers, improving expressivity while preserving symplectic structure by construction. We further introduce a universal approximation theorem covering \textsc{GRB-SympNet}'s approximation capacity (see Appendix~\ref{app:grb_sympnet_uat}).

\item \textbf{Cross system evaluation.}
We evaluate \textsc{CaLiSym} on a simulated, forced and dissipative double pendulum, a \textbf{real-world} aerial quadrotor and a \textbf{real-world} contact-constrained quadruped. 
\end{itemize}

To the best of our knowledge, this is the first physics informed symplectic learning framework demonstrated across \textit{real-world robotic systems} whose observed dynamics are controlled, dissipative, or contact constrained. The results show that the proposed framework is compatible in principle with any exactly symplectic predictor, scales across system dimensions, and improves long-horizon autoregressive prediction while remaining parameter efficient.

\vspace{-2mm}
\section{Related Work}
\textsc{CaLiSym} builds upon several developments spanning robot dynamics learning~\cite{vaidhyanathan2025metasym}, geometric machine learning~\cite{geometric21bronstein}, and structure preserving representations of physical systems~\cite{structure21hernandez}. We review the works most closely related to the proposed framework and organize them according to the key design principles underlying our approach.
\vspace{-2mm}
\subsection{Learning Dynamics Models}
Learning dynamics from data is a fundamental problem in robotics, underpinning model predictive control~\cite{mastalli2022agile}, state estimation~\cite{martinez25estimation}, system identification~\cite{martinez2026identificationconstraintsdisturbancebayesian}, and model-based reinforcement learning~\cite{mastalli2022agile,barfoot2024state,moerland2022mbrl}. Existing approaches span a broad spectrum of learning paradigms. At one end, black-box neural predictors and sequence models directly learn state transitions from data and have demonstrated strong empirical performance across a variety of robotic systems~\cite{ha2018worldmodels,hafner2020dreamer,sanchez2020learning}. More recently, learned world models have extended this paradigm by constructing latent representations that jointly capture dynamics, observations, and control, enabling planning and imagination-based reinforcement learning~\cite{ha2018worldmodels,hafner2020dreamer}. However, because these models are primarily black-box, they generally provide no guarantees regarding the preservation of physical quantities such as energy, momentum, or geometric structure, often leading to drift and instability during long-horizon rollouts~\cite{karoly2020deep}.

A complementary direction is operator learning, where the objective is to learn mappings between function spaces rather than individual trajectories. Architectures such as Fourier Neural Operators (FNOs)~\cite{li2020fourier,li2023fourier} and related neural operators have achieved remarkable success in modeling complex physical systems and partial differential equations by exploiting global spectral representations. While highly expressive, these methods are fundamentally designed to approximate solution operators and likewise do not explicitly enforce the geometric structure underlying mechanical systems. As a result, conservation laws and phase-space invariants may be violated even when short-term prediction accuracy is high, while their scalability is yet to be proven.

Within robotics, specialized sequence architectures such as RWM~\cite{li2025roboticworldmodelneural} augment autoregressive (AR) prediction with latent memory representations to improve long-horizon forecasting and capture history-dependent effects. By maintaining a learned latent state, these models can represent complex non-Markovian behaviors and partially mitigate error accumulation over extended prediction horizons. Nevertheless, the learned latent dynamics remain largely unconstrained by the underlying physical laws, providing no guarantees regarding the preservation of conservation laws, geometric structure, or physically meaningful invariants. As a result, predictions may become physically inconsistent under distribution shift or during long-horizon rollouts. Furthermore, training such recurrent and autoregressive world models typically relies on backpropagation through time (BPTT)~\cite{bptt90werbos,pascanu13rnn}, requiring gradients to be propagated through long computational graphs. This introduces substantial memory and computational overhead while also exposing training to the well-known challenges of vanishing and exploding gradients, which can negatively impact optimization stability and scalability to increasingly long prediction horizons.

The common limitation across these families of methods is that physical structure is treated as an emergent property to be learned from data rather than a fundamental constraint imposed by the model architecture. In contrast, \textsc{CaLiSym} incorporates a strong geometric inductive bias by constructing the learned dynamics directly within a symplectic framework. This enables the model to preserve the underlying phase-space structure by design, substantially improving long-horizon stability, robustness to distribution shift, and the faithful reproduction of physically meaningful invariants. Its training relies entirely on teacher forcing, while the \textsc{CaLiSym}'s predictions are entirely Markovian (i.e. single-step maps).

\vspace{-2mm}
\subsection{Geometric Models of Physical Systems}
A complementary line of research seeks to improve generalization and long-horizon stability by embedding physical structure directly into the learning architecture rather than relying solely on data-driven regularization~\cite{karniadakis2021physics}. Hamiltonian Neural Networks (HNNs) learn a Hamiltonian function whose gradients generate the system dynamics~\cite{greydanus2019hamiltonian}, while Lagrangian Neural Networks (LNNs) parameterize the Lagrangian and recover the equations of motion through the Euler--Lagrange formalism~\cite{cranmer2020lagrangian}. These approaches leverage fundamental principles of classical mechanics to restrict the learned dynamics to physically plausible trajectories, resulting in improved extrapolation and data efficiency compared to unconstrained neural predictors.

A related but distinct class of methods focuses on preserving geometric structure directly at the level of the flow map. Symplectic neural networks such as \textsc{SympNet}s~\cite{jin2020sympnets} construct discrete-time updates that preserve the canonical symplectic form exactly. By learning the dynamics map itself rather than an underlying energy function, these architectures avoid repeated differentiation during rollout and inherit the long-term stability properties associated with geometric numerical integrators. Consequently, they have emerged as one of the most promising directions for learning stable dynamics models from data, especially in data-scarce regimes.

Despite these advances, existing geometric learning methods remain largely confined to conservative mechanical systems. Real robotic platforms are fundamentally non-conservative, since they exchange energy with their environment through actuation, damping, friction, impacts, contacts, sensing, and control. As a result, the assumptions underlying the classical Hamiltonian formalism are frequently violated in practice. While HNNs and LNNs can incorporate certain extensions, they still require recovering the dynamics through differentiating learned energy functions, increasing computational cost and sensitivity to approximation errors. Existing symplectic architectures, on the other hand, preserve geometric structure exactly but are generally unable to represent dissipative or externally driven dynamics without sacrificing their underlying theoretical guarantees.

\textsc{CaLiSym} bridges this gap by extending structure-preserving dynamics learning \textit{beyond} conservative systems. Through a lifted symplectic representation, it retains the favorable stability and geometric properties of symplectic models while accommodating the strongly dissipative, actuated, and constrained dynamics encountered in real-world robotics. Moreover, the resulting predictor remains lightweight, parameter-efficient, and requires only a single forward pass per time step, making it practical for large-scale robotic learning and control applications.

\vspace{-2mm}
\subsection{Open and Constrained Dynamical Systems}
Most real-world robotic systems are fundamentally non-conservative. Through actuation, damping, friction, impacts, contacts, and feedback control, they continuously exchange energy, momentum, and information with their environment. To model such phenomena, geometric mechanics has developed a variety of extensions to classical Hamiltonian theory, including port-Hamiltonian systems, dissipative and metriplectic dynamics, and controlled mechanical systems, which incorporate energy exchange and irreversible processes while preserving an underlying geometric structure~\cite{vanderschaft2017l2gain,maschke92port,morrison84bracket}. Likewise, constrained Hamiltonian mechanics, presymplectic geometry, and Dirac structures provide principled frameworks for representing contacts, kinematic constraints, and admissible motions~\cite{yoshimura2006diracSI}. 

A recurring theme across these formulations is that apparently non-Hamiltonian behavior can often be represented as Hamiltonian dynamics on an appropriately augmented state space. Symplectification, cotangent lifts, and extended phase-space constructions introduce auxiliary coordinates through which forcing, dissipation, and constraints arise as projections of a higher-dimensional conservative system~\cite{abraham2008foundations,betancourt2018symplectic}. Despite their importance in geometric mechanics, these ideas have seen limited adoption in dynamics learning, where most symplectic architectures remain restricted to conservative systems. A recent exception learns the symplectification itself through Dirac-structure gauge fixing, representing non-conservative effects through a scalar Rayleigh-type dissipation field~\cite{papatheodorou2026symplectification}. In contrast, \textsc{CaLiSym} conditions the lifted symplectic dynamics directly on the measured actuation and contact wrenches as explicit ports, without assuming that the underlying interactions admit a scalar potential. This extends symplectic learning to controlled, dissipative, and contact-constrained robotic systems while retaining the structure-preserving properties of the lifted symplectic representation.

\subsection{Local Function Approximation}
Recent advances in neural approximation theory have highlighted the benefits of localized function representations. Spline-based models and Kolmogorov--Arnold Networks (KANs) exploit the observation that many high-dimensional nonlinear mappings can be represented efficiently through compositions of low-dimensional functions, often yielding superior parameter efficiency and interpretability compared to conventional multilayer perceptrons~\cite{igelnik2003kolmogorov,liu2024kan}. In particular, local basis functions are well suited to physical systems whose dynamics exhibit strongly nonlinear behavior only within restricted regions of the state space, a common characteristic of robotic systems with contacts, saturation effects, and configuration-dependent dynamics.

Despite their approximation power, generic spline and KAN-style architectures are purely function approximators and do not preserve the geometric structure of the underlying dynamical system. Consequently, improved local accuracy does not necessarily translate into physically consistent long-horizon predictions. For dynamical systems, preserving the structure of the flow can be as important as accurately approximating the vector field itself.

Our proposed \textsc{GRB-SympNet} combines the advantages of local function approximation with exact geometric preservation (see Appendix~\ref{app:grb_sympnet_uat}). Building upon the generalized ridge formulation of SympNets, it replaces global nonlinearities with compact tensor-product spline representations, enabling highly expressive local approximations of the scalar Hamiltonian while preserving symplecticity exactly by construction. The resulting architecture inherits the parameter efficiency and locality of spline-based models while retaining the long-term stability and geometric guarantees of structure-preserving symplectic networks.

\vspace{-3mm}
\section{Robot Dynamics and Hamiltonian Mechanics}
Robot dynamics is primarily described by the Equations of Motion (EoMs) that connect the forces acting on the rigid bodies comprising the system and the accelerations they produce. There has been extensive research~\cite{featherstone2014rigid} on algorithms that efficiently assemble these EoMs, with most notable the Recursive Newton-Euler Algorithm (RNEA) and the Articulated-Body Algorithm (ABA)~\cite{buondonno2015recursive,featherstone1983calculation}. However, they all rely on the rigid-body assumption, while phenomena such as friction are commonly modeled using the Lagrangian formalism.

\vspace{-3mm}
\subsection{Lagrangian Formalism}
The Lagrangian formalism provides the variational counterpart to the Newton--Euler bookkeeping that RNEA and ABA carry out~\cite{goldstein2001classical}. A robot's configuration is a point on a smooth manifold rather than in a vector space, with the manifold determined by its kinematic structure. For a free-floating base together with $n$ revolute joints, the base pose belongs to the special Euclidean group $\mathbb{SE}(3)$ and the joint angles to $\mathbb R^{n}$, so the configuration is the composite Lie group
\begin{equation}
\mathcal Q = \mathbb{SE}(3)\times\mathbb R^{n},
\end{equation}
a manifold of dimension $6+n$. Each configuration $\mathbf q\in\mathcal Q$ carries velocities in its tangent space $T_{\mathbf q}\mathcal Q$, which the Lie-group structure identifies with the Lie algebra $\mathfrak{se}(3)\times\mathbb R^{n}\cong\mathbb R^{6+n}$. The generalized velocity $\boldsymbol v$ pairs the base twist with the joint rates, the base component being an element of $\mathfrak{se}(3)$ rather than a derivative of pose coordinates. It relates to the configuration rate $\dot{\mathbf q}\in T_{\mathbf q}\mathcal Q$ through the tangent map of left translation,
\begin{equation}
    \dot{\mathbf q} = \mathbf q\,\boldsymbol v^{\wedge},
\end{equation}
where $(\cdot)^{\wedge}$ maps the twist to its matrix form in $\mathfrak{se}(3)$~\cite{sola2021micro}. On the Euclidean factor this reduces to the identity $\dot{\mathbf q}=\boldsymbol v$, while on $\mathbb{SE}(3)$ it reconstructs the pose derivative from the body twist. The state $(\mathbf q,\boldsymbol v)$ then lives in the tangent bundle $T\mathcal Q$.

On this bundle the Lagrangian is a map $L:T\mathcal Q\to\mathbb R$ given by kinetic minus potential energy,
\begin{equation}
L(\mathbf q,\boldsymbol v):=T(\mathbf q,\boldsymbol v)-V(\mathbf q),
\end{equation}
where $T(\mathbf q,\boldsymbol v)$ is the kinetic energy and $V(\mathbf q)$ the potential energy. The Euler--Lagrange equations on $\mathcal Q$ yield the canonical manipulator equation,
\begin{equation}
\mathbf M(\mathbf q)\,\dot{\boldsymbol v}+\mathbf h(\mathbf q,\boldsymbol v)+\mathbf g(\mathbf q)=\boldsymbol\tau,
\end{equation}
with $\mathbf M(\mathbf q)$ the generalized mass matrix, $\mathbf h(\mathbf q,\boldsymbol v)$ the Coriolis and centrifugal terms, $\mathbf g(\mathbf q)$ the gravitational generalized force, and $\boldsymbol\tau$ the applied generalized forces.

While the Lagrangian formulation on the tangent bundle $T\mathcal{Q}$ provides a familiar framework for deriving a robot's equations of motion, its velocity-based state $(\mathbf{q},\boldsymbol v)$ carries three structural limitations. First, naive integration of the resulting vector field preserves none of the geometric invariants of the flow, so long-horizon simulations accumulate artificial energy drift and violate phase-space volume preservation. Second, symmetries and their associated conserved quantities remain implicit, rather than appearing as explicit geometric objects that integrators and learning models can exploit. Third, algebraic constraints from intermittent contact or closed kinematic loops are typically handled by penalty terms or Lagrange multipliers appended to the dynamics, an approach that degrades when the velocity-to-momentum map $\mathbf{p}=\mathbf{M}(\mathbf{q})\boldsymbol v$ becomes ill-conditioned under constraint projection and destabilizes the recursion.
\vspace{-3mm}
\subsection{Hamiltonian Formalism}
Transitioning to the Hamiltonian formulation on the cotangent bundle $T^*\mathcal{Q}$, which pairs configurations with generalized momenta rather than velocities, resolves several limitations of the classical Lagrangian framework. The cotangent bundle is naturally equipped with the canonical symplectic two-form
\begin{equation}
    \omega = \sum_{i=1}^{n} dq^i \wedge dp_i,
\end{equation}
allowing the dynamics to be expressed as the first-order Hamiltonian system
\begin{equation}
    \dot{\mathbf q}=\frac{\partial H}{\partial \mathbf p},
    \qquad
    \dot{\mathbf p}=-\frac{\partial H}{\partial \mathbf q},
\end{equation}
where $H$ denotes the Hamiltonian of the system.

Beyond providing an equivalent description of the dynamics, this formulation exposes the geometric structure underlying mechanical systems. The Hamiltonian flow preserves the symplectic form exactly, and consequently preserves phase-space volume through Liouville's theorem. When discretized using symplectic integrators, this structure often leads to favorable long-horizon numerical behavior and bounded energy drift. Furthermore, Noether's theorem associates continuous symmetries with conserved momentum maps $\mathbf J:T^*\mathcal Q\rightarrow\mathfrak g^*$, exposing physically meaningful invariants such as linear and angular momentum. Finally, constrained Hamiltonian mechanics and the associated Dirac bracket formalism provide a principled mechanism for restricting trajectories to admissible constraint manifolds, making contacts and kinematic constraints part of the geometric structure of the system rather than external corrections.

These properties are particularly attractive from a learning perspective. By restricting the admissible dynamics to symplectic transformations on $T^*\mathcal Q$, the hypothesis space is dramatically reduced, providing a powerful geometric inductive bias that can improve data efficiency, extrapolation, and long-horizon stability.

\vspace{-4mm}
\subsection{Symplecticity in Machine Learning}
Beyond their physical interpretation, these geometric properties fundamentally reshape the learning problem. By restricting the admissible dynamics to symplectomorphisms on $T^*\mathcal{Q}$, the model no longer needs to approximate arbitrary state transitions, but only those consistent with the underlying mechanics~\cite{guillemin1990symplectic, weinstein1977lectures}. This restriction acts as a strong inductive bias, reducing the effective hypothesis space and improving long horizon prediction, where unconstrained neural models often accumulate geometric errors that manifest as energy drift, instability, and violation of conserved quantities.

Since the Hamiltonian flow preserves the symplectic form exactly, satisfying $\mathcal{L}_{\mathbf{X}_H}\omega = 0$~\cite{koszul2019introduction}, a data driven model can be constrained to learn a discrete time transition map $\Phi_\theta$ satisfying $\Phi_\theta^*\omega=\omega$. Embedding this invariant directly into the network architecture yields learned dynamics that inherit the geometric structure of the underlying system, resulting in minimal long horizon energy error, volume preserving evolution, and improved autoregressive stability.

This perspective has motivated a broad class of Hamiltonian and symplectic neural networks. Their strongest theoretical guarantees, however, apply only when the observed state evolves as a closed conservative Hamiltonian system. Most robotic systems do not satisfy this assumption. A torque controlled double pendulum exchanges energy through actuation and damping. A quadrotor is driven by thrust and aerodynamic forces, while a legged robot exchanges energy and impulses through intermittent contact with the environment. Consequently, the measured physical state $\mathbf{x} := (\mathbf{q},\mathbf{p})$ is generally forced, dissipative, and contact constrained. Imposing symplecticity directly on this state therefore enforces the wrong geometric structure.

\vspace{-4mm}
\subsection{Symplectic Learning beyond Conservative Systems}

This paper proposes \textsc{CaLiSym}. Rather than enforcing symplecticity on the measured physical state, we impose it on a structured lifted phase space that augments the state with variables governing energy exchange with the environment~\cite{betancourt2018symplectic}. Dynamics are learned using exactly symplectic neural predictors in this lifted space and subsequently projected back to the physical state. Consequently, the learned model preserves symplectic structure in the lifted representation while remaining compatible with actuation, dissipation, and contact in the observed dynamics.

The lift is structured rather than latent. Given a physical state and port variables $\mathbf e$, such as control inputs or contact forces, the system is embedded into a gauge fixed section of a canonical lifted phase space. Prediction proceeds through lifting, symplectic evolution, projection, and re-embedding,
\begin{equation}
  \mathbf x_k \xrightarrow{\;\sigma_{\mathbf e_k}\;} \mathbf z_k
       \xrightarrow{\;\Phi_\theta\;} \bar{\mathbf z}_{k+1}
       \xrightarrow{\;\Pi_{\mathcal X}\;} \hat{\mathbf x}_{k+1}.
\end{equation}
For autoregressive rollout, the projected state is lifted again using the next port value,
\begin{equation}
  \mathbf z_{k+1} = \sigma_{\mathbf e_{k+1}}(\hat{\mathbf x}_{k+1}).
\end{equation}
This construction enables a single symplectic predictor to model systems with time varying controls, dissipation, and intermittent contact while preserving exact symplectic structure in the lifted space.

We also introduce \textbf{\textsc{GRB-SympNet}}, a generalized-ridge B-spline SympNet. \textsc{GRB-SympNet} keeps the exact symplectic generalized-ridge layer, but parameterizes the scalar ridge Hamiltonian with a compact B-spline/KAN-style local function. This improves local expressivity in low-dimensional lifted systems while preserving exact symplecticity. In this paper, \textsc{GRB-SympNet} is used for the controlled dissipative double pendulum. For the higher-dimensional quadrotor and quadruped experiments, we use \textsc{GR-SympNet} as the scalable lifted predictor. This separation is intentional: the structured lift is the transferable contribution across systems, while \textsc{GRB-SympNet} is an architecture contribution evaluated in the low-dimensional setting where spline ridge parameterizations are most appropriate.

We evaluate \textsc{CaLiSym} on three systems of increasing complexity. The controlled dissipative double pendulum isolates the effect of the lift and tests whether \textsc{GRB-SympNet} improves low-dimensional lifted symplectic prediction. The quadrotor tests whether the same lifted formulation handles unbounded forced dynamics with time-varying controls. The quadruped tests whether the structured lift scales to high-dimensional contact-rich robot dynamics when paired with \textsc{GR-SympNet}. Together, these experiments test the main claim of the paper: symplectic robot learning can be extended beyond conservative systems by lifting the physical dynamics into a structured canonical phase space.

\section{Methods}
\label{sec:methods}

The goal of \textsc{CaLiSym} is to learn a discrete time predictor for robotic systems whose measured physical dynamics are not closed Hamiltonian systems. The key idea is to separate where symplecticity is enforced from where prediction is evaluated. The measured state and its interaction variables are embedded into a structured canonical phase space, an exactly symplectic map is learned in that space, and the physical trajectory is recovered by projection. This section formalizes the lifted representation, states the geometric properties of the construction, describes the symplectic predictors used in our experiments, and gives the training, normalization, and verification procedures.
\vspace{-3mm}
\subsection{Problem Setting}

Consider a robotic system evolving on a configuration manifold $\mathcal Q$ with generalized coordinates $\mathbf q \in \mathcal Q$ and generalized momenta,
\begin{equation}
    \mathbf p = \mathbf M(\mathbf q)\boldsymbol{v},
\end{equation}
where $\mathbf M(\mathbf q)$ denotes the generalized mass matrix and $\boldsymbol v \in T_{\mathbf q}\mathcal Q$ the generalized velocity.
The measured physical state is therefore
\begin{equation}
    \mathbf x = (\mathbf q,\mathbf p) \in T^*\mathcal Q.
\end{equation}

Unlike closed Hamiltonian systems, robotic systems continuously exchange energy and momentum with their environment through actuation, dissipation, and contact. We represent these interactions through a collection of \emph{physical ports},
\begin{equation}
    \mathbf e \in \mathcal E,
\end{equation}
whose precise interpretation depends on the system under consideration. Examples include actuator torques, rotor thrusts, contact forces, or other externally supplied interaction variables.

A broad class of robotic systems with constraints such as contacts can be written as,
\begin{equation}
\begin{aligned}
    &\mathbf M(\mathbf q)\dot{\boldsymbol v} + \mathbf h(\mathbf q, \boldsymbol v) + \mathbf g(\mathbf q) + \mathbf c_{\mathrm{diss}}(\boldsymbol{v}) = \mathbf B(\mathbf q)\mathbf u + \mathbf J_c(\mathbf q)^\top \mathbf f_c\\
    &\quad\text{s.t.: } \phi(\mathbf q)=0, \text{ with } \mathbf J_c(\mathbf q)=\frac{\partial\phi}{\partial{\mathbf q}}.
\end{aligned}
\end{equation}
The corresponding energy balance is
\begin{equation}
\frac{dH}{dt}
=
\boldsymbol v^\top \mathbf B(\mathbf q)\mathbf u
+
\boldsymbol v^\top \mathbf J_c(\mathbf q)^\top\mathbf f_c
-
\boldsymbol v^\top \mathbf c_{\mathrm{diss}}(\boldsymbol v),
\end{equation}
where
\[
H(\mathbf q,\boldsymbol v)
=
\frac12\boldsymbol v^\top\mathbf M(\mathbf q)\boldsymbol v
+
V(\mathbf q).
\]
For ideal bilateral constraints satisfying $\phi(\mathbf q)=0$, differentiation gives $\mathbf J_c(\mathbf q)\boldsymbol v=0$, so the contact power vanishes and the constraints perform no work. Real robotic contact, however, is typically unilateral, frictional, and impact-driven, introducing energy exchange and discontinuous momentum updates that lie outside the smooth balance above. Consequently, the measured state generally does not evolve as a closed conservative Hamiltonian system. The central challenge is therefore to retain the geometric guarantees of symplectic learning while modeling dynamics that are inherently open. \textsc{CaLiSym} addresses this challenge by constructing a lifted representation that admits an exactly symplectic formulation while remaining faithful to the observed physical dynamics.

\subsection{Structured Lift and Augmented Cotangent-lifted space}
\label{subsec:structured_lift_rollout}

The central idea of \textsc{CaLiSym} is to enforce symplecticity on an augmented canonical phase space rather than on the measured physical state itself. Let
\begin{equation}
    \boldsymbol{x}
    =
    (\boldsymbol{q},\boldsymbol{p})
    \in
    \mathcal{X}
    :=
    T^*\mathcal{Q}
\end{equation}
denote the measured physical state, and let $\boldsymbol e\in\mathcal E$ denote the physical ports through which the system exchanges energy and momentum with its environment. Since the physical subsystem is generally open, the transition map on $\mathcal X$ need not be symplectic. We therefore embed the physical state and ports into a structured lifted phase space on which an exactly symplectic map can be learned.

We define the lifted canonical state,
\begin{equation}
    \boldsymbol{Z}
    =
    (\boldsymbol{Q},\boldsymbol{P})
    \in
    \mathcal{Z},
    \qquad
    \Omega
    =
    \sum_i dQ^i\wedge dP_i ,
\end{equation}
where $\Omega$ is the canonical symplectic form in the augmented space. For a controlled system with contact ports, we use:
\begin{align}
    \boldsymbol{Q}
    &=
    (\boldsymbol{q},\boldsymbol{y},
    \boldsymbol{\lambda}_u,\boldsymbol{\lambda}_c),
    \\
    \boldsymbol{P}
    &=
    (\boldsymbol{r},\boldsymbol{p},
    \boldsymbol{\mu}_u,\boldsymbol{\pi}_c).
    \label{eq:canonical_lift_layout}
\end{align}
Here $(\boldsymbol{q},\boldsymbol{p})$ are the physical configuration and momentum variables. The pair $(\boldsymbol{r},\boldsymbol{y})$ consists of auxiliary conjugate fibers associated with the physical state. The pair $(\boldsymbol{\lambda}_u,\boldsymbol{\mu}_u)$ represents the actuation port, and $(\boldsymbol{\lambda}_c,\boldsymbol{\pi}_c)$ represents the constraint port. The lift is generic, hence blocks that are not present for a given system are omitted. With this ordering, the lifted symplectic form is
\begin{equation}
\begin{aligned}
    \Omega &= d\boldsymbol{q}\wedge d\boldsymbol{r}
    +
    d\boldsymbol{y}\wedge d\boldsymbol{p}
    +
    d\boldsymbol{\lambda}_u\wedge d\boldsymbol{\mu}_u
    +
    d\boldsymbol{\lambda}_c\wedge d\boldsymbol{\pi}_c .
\end{aligned}
\end{equation}
Here wedge products between vector-valued variables denote the corresponding componentwise sum. Equivalently, using a symplectic rotation transformation we define the equivalent base--fiber coordinates,
\begin{align}
    \boldsymbol{b}
    &=
    (\boldsymbol{q},\boldsymbol{p},
    \boldsymbol{\mu}_u,\boldsymbol{\lambda}_c),
    \\
    \boldsymbol{\zeta}
    &=
    (\boldsymbol{r},-\boldsymbol{y},
    -\boldsymbol{\lambda}_u,\boldsymbol{\pi}_c).
    \label{eq:base_fiber_coordinates}
\end{align}
Then the symplectic form becomes,
\begin{equation}
    \Omega
    =
    \sum_i db^i\wedge d\zeta_i .
\end{equation}
Thus, the lifted variables form a canonical cotangent representation over the base variables $\boldsymbol b$. The purpose of the lift is not merely to augment the state dimension, but to transform the learning problem into one of modeling Hamiltonian dynamics on an augmented canonical symplectic manifold. The lifted dynamics are generated by a Hamiltonian on this augmented phase space. Let
\begin{equation}
\boldsymbol a(\boldsymbol q,\boldsymbol p)
\end{equation}
denote the physical configuration dynamics and let
\begin{equation}
\boldsymbol f(\boldsymbol q,\boldsymbol p,\boldsymbol\mu_u,\boldsymbol\lambda_c)
\end{equation}
denote the corresponding momentum dynamics, including the effects of actuation, contact forces, and dissipation. Motivated by the cotangent lift of a forced vector field, we consider lifted Hamiltonians of the form,
\begin{equation}
\begin{aligned}
    \mathcal H_{\mathrm{lift}}(\mathbf b,\boldsymbol \zeta) = \boldsymbol r^\top \boldsymbol a(\boldsymbol q,\boldsymbol p) 
        - \boldsymbol y^\top \boldsymbol f(\boldsymbol q, \boldsymbol p, \boldsymbol\mu_u, \boldsymbol\lambda_c) + G(\boldsymbol b),
\end{aligned}
\label{eq:lifted_hamiltonian}
\end{equation}
where $G$ is an arbitrary scalar function of the base variables. The first term propagates the configuration dynamics, while the second couples the physical momentum evolution to the auxiliary fibers. The Hamiltonian is linear in the fiber variables $(\boldsymbol r,\boldsymbol y)$ and therefore represents a cotangent lift of the physical dynamics augmented with canonical port variables.

\subsection{Dynamics and Section Invariance}

The lifted Hamiltonian induces a coupled evolution of the physical and auxiliary variables. The physical subsystem evolves according to:
\begin{equation}
\begin{aligned}
\dot{\mathbf{q}} &= \boldsymbol a(\mathbf q,\mathbf p), \quad \dot{\mathbf p} = \mathbf f(\mathbf q, \mathbf p, \boldsymbol \mu_u, \boldsymbol{\lambda}_c), \\
\dot{\mathbf r} &= -(\partial_{\mathbf q} \boldsymbol a)^\top \mathbf r + (\partial_{\mathbf q} \mathbf f)^\top \mathbf y - \nabla_{\mathbf q} G,\\
\dot{\mathbf y} &= (\partial_{\mathbf p} \boldsymbol a)^\top \mathbf r - (\partial_{\mathbf p} \mathbf f)^\top \mathbf y + \nabla_{\mathbf p} G, \\
\dot{\boldsymbol \mu_u} &= \mathbf 0,\quad \dot{\boldsymbol \lambda}_u = -(\partial_{\boldsymbol{\mu_u}} \mathbf f)^\top \mathbf y + \nabla_{\boldsymbol \mu_u} G,\\
\dot{\boldsymbol \lambda}_c &= \mathbf 0,\quad \dot{\boldsymbol \pi}_c = (\partial_{\boldsymbol{\lambda}_c} \mathbf f)^\top \mathbf y - \nabla_{\boldsymbol \lambda_c} G,
\end{aligned}
\end{equation}

The lifted dynamics evolve on the full augmented phase space $\mathcal Z$. In practice, however, only the physical state and interaction ports are observed. The measured data therefore occupy a distinguished gauge-fixed section of the lifted space obtained by fixing the port variables and setting all auxiliary fibers to zero.

Under this gauge choice and given the ports,
\begin{equation}
    \boldsymbol e_k = (\boldsymbol u_k,\boldsymbol f_{c,k}),
\end{equation}
we define the physical data section,
\begin{equation}
\begin{split}
\mathcal{S}_{\boldsymbol{e}_k}
=
\Big\{
\boldsymbol{Z}\in\mathcal{Z}:\;&
\boldsymbol{r}=\boldsymbol{0},\;
\boldsymbol{y}=\boldsymbol{0},\;
\boldsymbol{\lambda}_u=\boldsymbol{0}, \\
&
\boldsymbol{\pi}_c=\boldsymbol{0},\;
\boldsymbol{\mu}_u=\boldsymbol{u}_k,\;
\boldsymbol{\lambda}_c=\boldsymbol{f}_{c,k}
\Big\}.
\end{split}
\label{eq:data_section}
\end{equation}

For systems without contacts, the contact block $(\boldsymbol{\lambda}_c,\boldsymbol{\pi}_c)$ is omitted. The corresponding section embedding
\begin{equation}
    \sigma_{\boldsymbol e_k} : \mathcal X \rightarrow \mathcal Z
\end{equation}
maps a measured state $\boldsymbol x_k=(\boldsymbol q_k,\boldsymbol p_k)$ onto $\mathcal S_{\boldsymbol e_k}$.

The physical section identifies the subset of lifted states that correspond to realizable observations. The physical coordinates and ports are preserved, while the auxiliary fibers vanish. Consequently, the physical dynamics are recovered by the projection $\Pi_{\mathcal X} : \mathcal Z \rightarrow \mathcal X$,
\begin{equation}
    \Pi_{\mathcal X}( \mathbf b,\boldsymbol \zeta) = (\boldsymbol q,\boldsymbol p).
\end{equation}

Although the physical section provides the initialization manifold for the observed data, it is not generally invariant under the lifted Hamiltonian dynamics. The auxiliary fibers may therefore evolve away from zero during prediction. This observation motivates the projection and re-embedding procedure introduced next, which restores consistency with the measured ports at every rollout step.

\subsection{Autoregressive Gauge-Fixed Lifted Rollout}
\label{subsec:ar_rollout}

The gauge-fixed section identifies the subset of the augmented phase space corresponding to physically realizable observations. Since the lifted Hamiltonian flow is not generally constrained to remain on this section, prediction proceeds by alternating between the physical section and the full lifted phase space. At each step, the current physical state is embedded into the augmented space, evolved by an exactly symplectic map, projected back to the physical state, and subsequently re-embedded using the next port value. 

A one-step prediction is computed as,
\begin{equation}
\begin{aligned}
    \boldsymbol{Z}_k
    &=
    \sigma_{\boldsymbol{e}_k}(\boldsymbol{x}_k), \\
    \bar{\boldsymbol{Z}}_{k+1}
    &=
    \Phi_\theta^{\Delta t}(\boldsymbol{Z}_k), \\
    \hat{\boldsymbol{x}}_{k+1}
    &=
    \Pi_{\mathcal{X}}(\bar{\boldsymbol{Z}}_{k+1}),
    \label{eq:one_step_lifted_prediction}
\end{aligned}
\end{equation}
where $\Phi_\theta^{\Delta t}:\mathcal{Z}\rightarrow\mathcal{Z}$ is exactly symplectic, i.e., $\left(\Phi_\theta^{\Delta t}\right)^*\Omega = \Omega$.
For autoregressive rollout, the projected physical prediction is re-embedded using the next measured or commanded port,
\begin{equation}
    \boldsymbol{Z}_{k+1}
    =
    \sigma_{\boldsymbol{e}_{k+1}}(\hat{\boldsymbol{x}}_{k+1}).
    \label{eq:regauge_step}
\end{equation}
Thus, the induced physical predictor is
\begin{equation}
    \hat{\boldsymbol{x}}_{k+1}
    =
    \Pi_{\mathcal{X}}
    \circ
    \Phi_\theta^{\Delta t}
    \circ
    \sigma_{\boldsymbol{e}_k}(\boldsymbol{x}_k),
    \label{eq:induced_physical_predictor}
\end{equation}
which need not be symplectic on the physical state alone.

\begin{algorithm}[t]
\caption{Gauge-Fixed Lifted Rollout}
\label{alg:lsd_rollout}
\KwIn{Initial state $\boldsymbol{x}_0$, ports $\{\boldsymbol{e}_k\}_{k=0}^{H}$, section embedding $\sigma$, projection $\Pi_{\mathcal{X}}$, symplectic predictor $\Phi_\theta^{\Delta t}$}
\KwOut{Predicted physical trajectory $\{\hat{\boldsymbol{x}}_k\}_{k=1}^{H}$}
$\boldsymbol{Z}_0 \leftarrow \sigma_{\boldsymbol{e}_0}(\boldsymbol{x}_0)$\;
\For{$k=0,\ldots,H-1$}{
    $\bar{\boldsymbol{Z}}_{k+1} \leftarrow \Phi_\theta^{\Delta t}(\boldsymbol{Z}_k)$\;
    $\hat{\boldsymbol{x}}_{k+1} \leftarrow \Pi_{\mathcal{X}}(\bar{\boldsymbol{Z}}_{k+1})$\;
    $\boldsymbol{Z}_{k+1} \leftarrow \sigma_{\boldsymbol{e}_{k+1}}(\hat{\boldsymbol{x}}_{k+1})$\;
}
\end{algorithm}

This lift--evolve--project--re-embed procedure separates the learned symplectic dynamics from the physical mechanisms responsible for energy exchange. The predictor evolves as an exactly symplectic map in the augmented phase space, while projection and re-embedding enforce consistency with the measured state and ports. Consequently, symplecticity is preserved in the lifted representation even though the induced dynamics on $\mathcal X$ may be controlled, dissipative, and contact constrained.

\subsection{Symplectic Predictors}
\label{subsec:symplectic_predictors}
The structured lift is architecture-agnostic, since any exactly symplectic map can be used as $\Phi_\theta^{\Delta t}$. In this paper, we use \textsc{GR-SympNet} for the quadrotor and quadruped experiments, and \textsc{GRB-SympNet} for the double pendulum. Both predictors are explicit discrete-time maps, so inference requires only a forward pass through the composed layers, not numerical integration of a learned vector field.
Let
\begin{equation}
    \mathbf{J}_{\Omega}
    =
    \begin{bmatrix}
        \mathbf{0} & \mathbf{I} \\
        -\mathbf{I} & \mathbf{0}
    \end{bmatrix}
\end{equation}
denote the canonical symplectic matrix. A differentiable map $\Phi:\mathcal{Z}\rightarrow\mathcal{Z}$ is symplectic when
\begin{equation}
    D\Phi(\boldsymbol{Z})^\top
    \mathbf{J}_{\Omega}
    D\Phi(\boldsymbol{Z})
    =
    \mathbf{J}_{\Omega}.
\end{equation}

\paragraph{\textsc{GR-SympNet}}
Let $(\boldsymbol{Q},\boldsymbol{P})\in\mathbb{R}^d\times\mathbb{R}^d$ denote lifted canonical coordinates. A \textsc{GR-SympNet} layer is generated by the scalar Hamiltonian
\begin{equation}
    \mathcal{H}_\ell(\boldsymbol{Q},\boldsymbol{P})
    =
    K_\ell(\boldsymbol{\xi}_\ell), \text{ where }
    \boldsymbol{\xi}_\ell
    =
    \mathbf{A}_\ell \boldsymbol{P}
    +
    \mathbf{B}_\ell \boldsymbol{Q},
    \label{eq:grsympnet_hamiltonian}
\end{equation}
and
\begin{equation}
    \mathbf{A}_\ell,\mathbf{B}_\ell\in\mathbb{R}^{m\times d} \text{ with }  m\ll d.
\end{equation}
Furthermore, symplecticity is preserved \textit{iff}
\begin{equation}
    \mathbf{A}_\ell \mathbf{B}_\ell^\top
    =
    \mathbf{B}_\ell \mathbf{A}_\ell^\top.
    \label{eq:isotropy_condition}
\end{equation}
Hamilton's equations for \eqref{eq:grsympnet_hamiltonian} give
\begin{align}
    \dot{\boldsymbol{Q}}
    &=
    \mathbf{A}_\ell^\top \nabla K_\ell(\boldsymbol{\xi}_\ell), \\
    \dot{\boldsymbol{P}}
    &=
    -
    \mathbf{B}_\ell^\top \nabla K_\ell(\boldsymbol{\xi}_\ell).
\end{align}
The ridge coordinate is invariant along the layer flow:
\begin{align}
    \dot{\boldsymbol{\xi}}_\ell
    &=
    \mathbf{A}_\ell \dot{\boldsymbol{P}}
    +
    \mathbf{B}_\ell \dot{\boldsymbol{Q}} \\
    &=
    \left(
    -
    \mathbf{A}_\ell \mathbf{B}_\ell^\top
    +
    \mathbf{B}_\ell \mathbf{A}_\ell^\top
    \right)
    \nabla K_\ell(\boldsymbol{\xi}_\ell) \\
    &=
    \boldsymbol{0}.
\end{align}
Therefore, the exact time-$h_\ell$ map is
\begin{align}
    \boldsymbol{Q}'
    &=
    \boldsymbol{Q}
    +
    h_\ell
    \mathbf{A}_\ell^\top
    \nabla K_\ell(\boldsymbol{\xi}_\ell), \\
    \boldsymbol{P}'
    &=
    \boldsymbol{P}
    -
    h_\ell
    \mathbf{B}_\ell^\top
    \nabla K_\ell(\boldsymbol{\xi}_\ell).
    \label{eq:grsympnet_layer}
\end{align}
Each layer is an exact Hamiltonian flow and is therefore symplectic. The full predictor is a composition
\begin{equation}
    \Phi_\theta^{\Delta t}
    =
    \phi_L\circ\cdots\circ\phi_1,
\end{equation}
and remains symplectic by closure of symplectic maps under composition. Since $\boldsymbol{\xi}_\ell\in\mathbb{R}^m$ with $m\ll d$, the nonlinear computation is performed in a low-dimensional ridge coordinate, which keeps the high-dimensional quadruped predictor parameter-efficient.

\paragraph{\textsc{GRB-SympNet}}
\textsc{GRB-SympNet} keeps the exact generalized-ridge symplectic layer in \eqref{eq:grsympnet_layer}, but changes the scalar function $K_\ell$. We write:
\begin{equation}
    K_\ell(\boldsymbol{\xi})
    =
    K_{\ell,\mathrm{spline}}(\boldsymbol{\xi})
    +
    K_{\ell,\mathrm{smooth}}(\boldsymbol{\xi}).
\end{equation}
The spline term is
\begin{equation}
    K_{\ell,\mathrm{spline}}(\boldsymbol{\xi})
    =
    \sum_{\alpha\in\mathcal{A}}
    c_{\ell,\alpha}
    \prod_{j=1}^{m}
    B_{\alpha_j}^{(r)}(\xi_j),
\end{equation}
where $B_{\alpha_j}^{(r)}$ is a degree-$r$ B-spline basis function. The smooth term provides a bounded global tail:
\begin{equation}
    K_{\ell,\mathrm{smooth}}(\boldsymbol{\xi})
    =
    \sum_{j=1}^{N_s}
    a_{\ell j}
    \varphi
    \left(
    \boldsymbol{w}_{\ell j}^\top \boldsymbol{\xi}
    +
    b_{\ell j}
    \right).
\end{equation}
Because $K_\ell$ appears only as a scalar Hamiltonian in the exact generalized-ridge layer, replacing it with a spline parameterization does not affect symplecticity. The spline is evaluated in the low-dimensional ridge coordinate $\boldsymbol{\xi}\in\mathbb{R}^m$, not in the full lifted state. This gives a local KAN-style approximation while preserving the explicit low-latency symplectic map. \textsc{GRB-SympNet}s are parameter efficient but they inherit the same limitation with KAN-based architectures in terms of scalability. Hence, we use them only for the double-pendulum experiments. Finally, similar to \eqref{eq:grsympnet_layer} the required gradients are calculated analytically, so no additional backward pass is required.

\subsection{Training, Normalization, and Verification}
\label{subsec:training_normalization_verification}

The datasets provide configurations and generalized velocities. We convert velocities to momenta using
\begin{equation}
    \boldsymbol{p}_k
    =
    \mathbf{M}(\boldsymbol{q}_k)\boldsymbol{v}_k.
\end{equation}
For the double pendulum and quadrotor, $\mathbf{M}(\boldsymbol{q})$ is computed from the analytical model. For the quadruped, $\mathbf{M}(\boldsymbol{q})$ is computed from the robot model. For systems with Lie-group configuration components, errors are computed in local coordinates. We write
\begin{equation}
    \boldsymbol{q}_1\ominus \boldsymbol{q}_2
    \in
    T_{\boldsymbol{q}_2}\mathcal{Q}
\end{equation}
for the local configuration difference. For Euclidean joints this is ordinary subtraction. For orientation components, it is computed using the logarithm map. The physical state error is
\begin{equation}
    \boldsymbol{x}_1\ominus \boldsymbol{x}_2
    =
    \left(
    \boldsymbol{q}_1\ominus \boldsymbol{q}_2,\;
    \boldsymbol{p}_1-\boldsymbol{p}_2
    \right).
\end{equation}
Hence, the one-step loss is
\begin{equation}
    \mathcal{L}_1 = \frac{1}{|\mathcal{D}|}\sum_k\left\|\mathbf{W}_x
    \left(\Pi_{\mathcal{X}}\left[
    \Phi_\theta^{\Delta t}(\sigma_{\boldsymbol{e}_k}\left(\boldsymbol{x}_k)\right)\right] \ominus \boldsymbol{x}_{k+1}\right) \right\|_2^2,
    \label{eq:one_step_loss}
\end{equation}
where $\mathbf{W}_x$ balances configuration and momentum errors.

We also penalize drift away from the data section before re-embedding. If
\begin{equation}
    \bar{\boldsymbol{Z}}_{j+1} = (\bar{\boldsymbol{b}}_{j+1},\bar{\boldsymbol{\zeta}}_{j+1})
\end{equation}
is the lifted prediction before gauge fixing, then
\begin{equation}
    \mathcal{L}_{\mathrm{sec}} =
    \frac{1}{|\mathcal{D}|}
    \sum_{k}
    \left\|
    \mathbf{W}_{\zeta}
    \bar{\boldsymbol{\zeta}}_{j+1}
    \right\|_2^2.
    \label{eq:section_loss}
\end{equation}
This term encourages the learned map to remain close to the measured data section over one step. It does not enforce symplecticity, since symplecticity is structurally enforced by the predictor architecture. The total objective is:
\begin{equation}
    \mathcal{L}
    =
    w_1\mathcal{L}_1
    +
    w_{\mathrm{sec}}\mathcal{L}_{\mathrm{sec}}.
    \label{eq:total_loss}
\end{equation}

The lifted variables have heterogeneous units and scales. We therefore normalize in base--fiber coordinates. For each pair $(b_i,\zeta_i)$, define
\begin{equation}
    \tilde b^i
    =
    \frac{b^i-\mu^i}{s^i},
    \qquad
    \tilde\zeta_i
    =
    s_i\zeta_i.
    \label{eq:symplectic_normalization}
\end{equation}
Then
\begin{equation}
    d\tilde b^i\wedge d\tilde\zeta_i
    =
    db^i\wedge d\zeta_i,
\end{equation}
so the normalization preserves the canonical symplectic form. This is essential, since independently standardizing all entries of $(\boldsymbol{b},\boldsymbol{\zeta})$ would generally destroy the canonical pairing.

Finally, although \textsc{GR-SympNet} and \textsc{GRB-SympNet} are symplectic by construction, we verify this numerically through the Jacobian residual
\begin{equation}
    \epsilon_{\mathbf{J}_{\Omega}}
    =
    \frac{
    \left\|
    D\Phi_\theta^{\Delta t}(\boldsymbol{Z})^\top
    \mathbf{J}_{\Omega}
    D\Phi_\theta^{\Delta t}(\boldsymbol{Z})
    -
    \mathbf{J}_{\Omega}
    \right\|_F
    }{
    \|\mathbf{J}_{\Omega}\|_F
    }.
    \label{eq:symplecticity_residual_app}
\end{equation}
For exactly symplectic predictors, this residual should be at numerical precision (see Appendix~\ref{app:symplecticity_conservation}). For unconstrained baselines, it is not expected to vanish. At inference time, the model is rolled out autoregressively using Algorithm~\ref{alg:lsd_rollout}. Each step amounts to a lift, one explicit symplectic forward pass, a projection, and a re-embedding with the next port value.

\section{Results}
\label{sec:results}
We evaluate \textsc{CaLiSym} on robotic systems whose observed dynamics violate the assumptions underlying classical symplectic learning. In particular, we consider controlled, dissipative, and contact-constrained systems, where the measured physical dynamics do not evolve as closed conservative Hamiltonian systems. The objective of these experiments is to determine whether enforcing symplectic structure in a structured lifted phase space improves long-horizon prediction while remaining applicable across systems of different dimensionality and complexity.

Our evaluation is designed to answer the following questions:
\begin{enumerate}
    \item Can symplectic learning be extended beyond conservative systems by enforcing symplectic structure in a structured lifted phase space?
    \item Does the proposed lifted formulation improve Markovian long-horizon autoregressive prediction while remaining parameter-efficient?
    \item Does the same lift construction transfer across robotic systems with substantially different dimensionality and interaction mechanisms, ranging from controlled dissipative systems to contact-rich locomotion?
\end{enumerate}

To investigate these questions, we consider three representative systems of increasing complexity. A controlled dissipative double pendulum serves as a low-dimensional benchmark for isolating the effect of the proposed lift and evaluating \textsc{GRB-SympNet} in the chaotic regime~\cite{strogatz94nonlinear}. A real-world quadrotor platform introduces floating-base, nonlinear  dynamics, testing whether the framework extends beyond simple mechanical systems. Finally, the real-world quadrupedal dynamics of ANYmal~D~\cite{hutter16anymal} involving floating-base motion, underactuation, and intermittent contact provides a demanding high-dimensional benchmark for evaluating scalability when combined with \textsc{GR-SympNet}. Together, these experiments assess the effectiveness of the proposed lifted symplectic representation across dissipative, controlled, and contact-constrained robotic systems, as well as its impact on long-horizon autoregressive prediction accuracy.

\begin{figure*}[ht]
  \centering
  \includegraphics[width=0.25\textwidth, trim=5 0 5 0, clip]{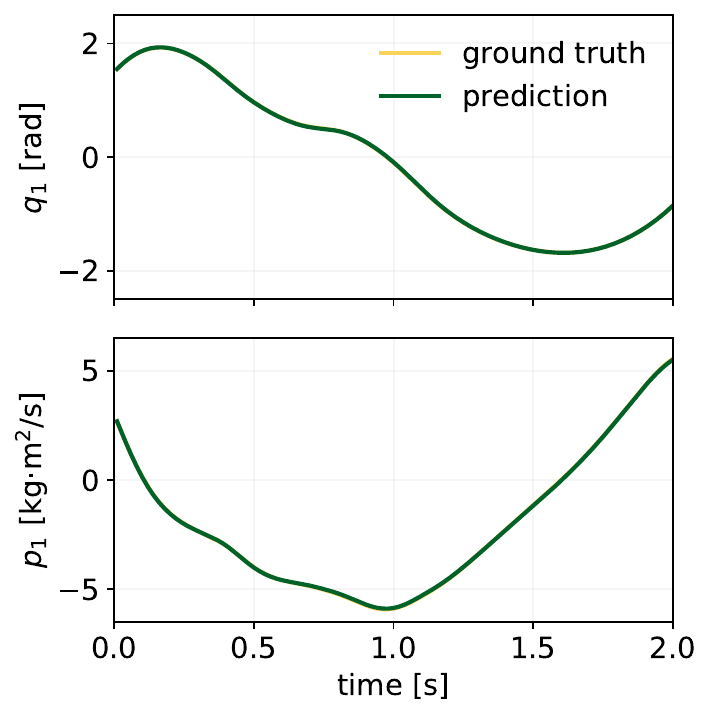}\hfill
  \includegraphics[width=0.25\textwidth, trim=5 0 5 0, clip]{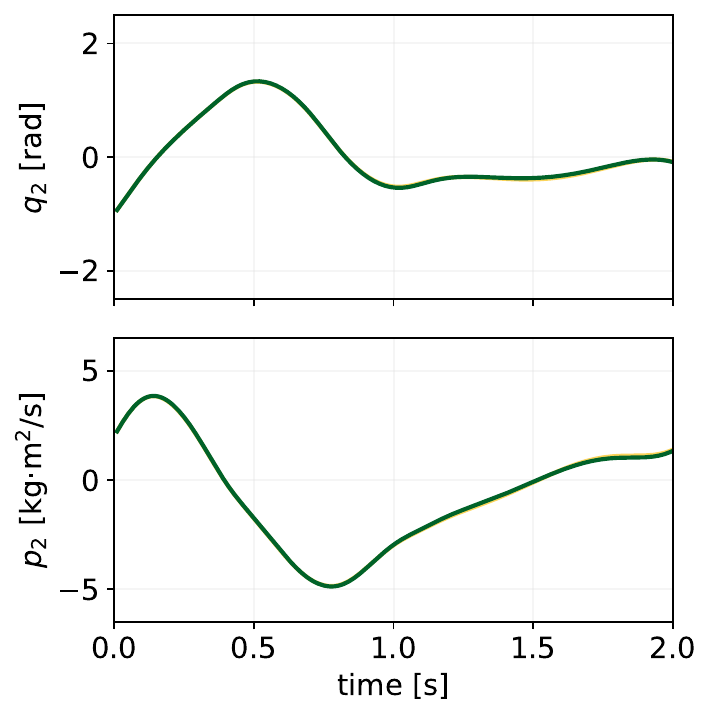}\hfill
  \includegraphics[width=0.5\textwidth, trim=-10 0 10 38, clip]{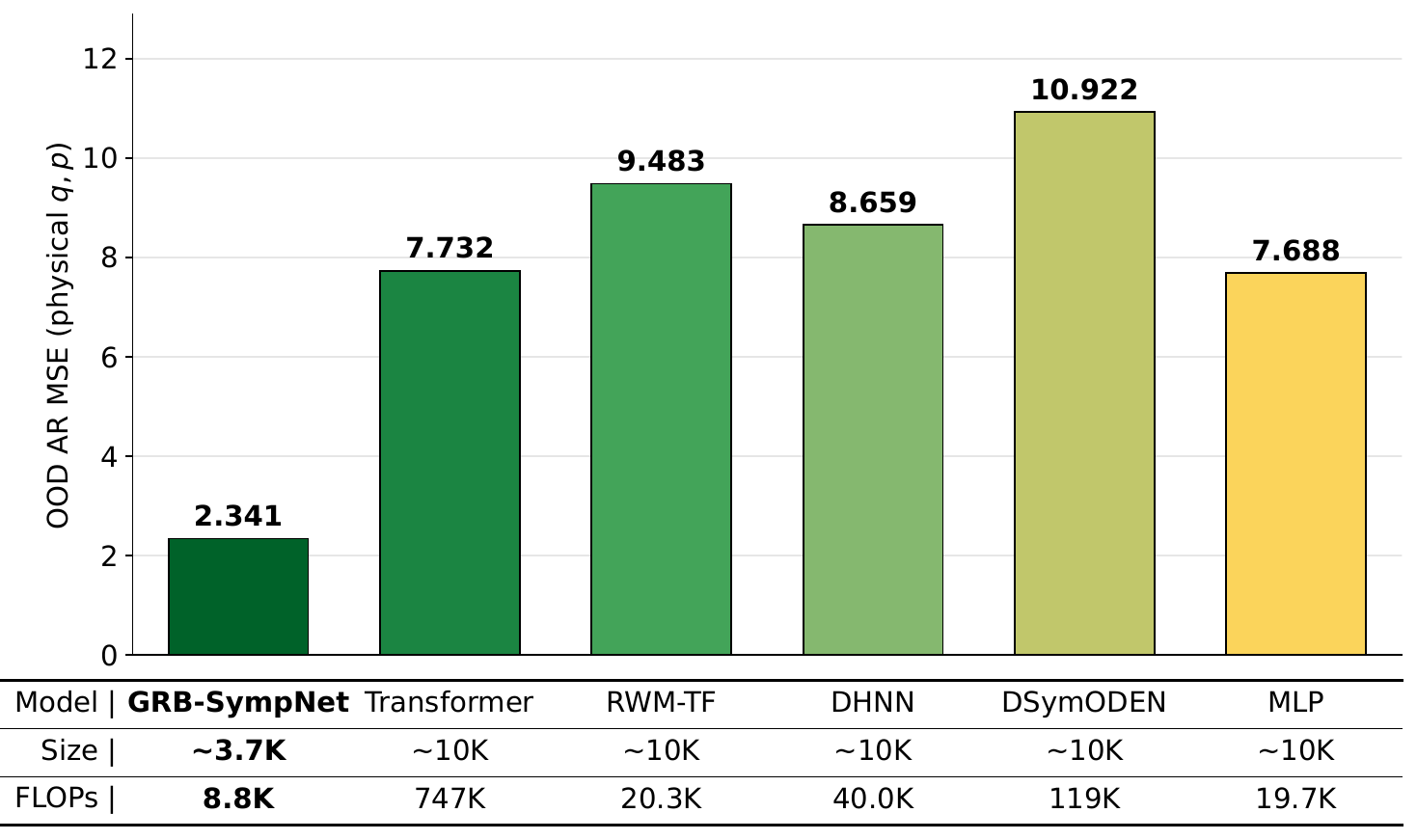}
  \vspace{-8mm}
  \caption{\textbf{Controlled dissipative double pendulum.} A simulated torque-controlled, damped system. (\textbf{Left}) An autoregressive out-of-distribution 200-timestep rollout of the physical state $(q_1,q_2,p_1,p_2)$, the joint angles and their corresponding momenta. (\textbf{Right}) The bar plot reports out-of-distribution autoregressive MSE and model size. \textsc{CaLiSym} with \textsc{GRB-SympNet} achieves the lowest OOD error ($2.341$ vs.\ $7.688$ for the strongest baseline, a $\sim$69.5\% reduction) using only $\sim$3.7K parameters against $\sim$10K for the baselines, showing the gain comes from structure rather than capacity. It is also the cheapest to evaluate, needing $8.8K$ FLOPs per step, under half the nearest baseline.}
  \label{fig:pendulum}
  \vspace{-5mm}
\end{figure*}

\subsection{Baselines and Metrics}
\label{subsec:baselines_metrics}
All models are trained using one-step teacher forcing and evaluated under autoregressive rollout (see Appendix~\ref{app:multi-horizon} for details). We deliberately avoid autoregressive training objectives to isolate the quality of the learned dynamics representation from the effects of the training procedure. While autoregressive training can improve rollout performance by exposing models to their own prediction errors, it may also encourage compensation for rollout-induced distribution shift and introduce additional optimization challenges for long prediction horizons~\cite{lamb2016professorforcingnewalgorithm,huang2025selfforcingbridgingtraintest}. Teacher forcing therefore provides a cleaner comparison of each architecture's ability to learn the underlying dynamics. For more details refer to Appendix~\ref{app:training_details}.

Importantly, all OOD evaluations are performed on trajectories drawn from previously unseen dynamical regimes, rather than on held-out segments of trajectories observed during training. The OOD splits correspond to distinct energy regimes for the double pendulum, held-out flight trajectories for the quadrotor, and unseen locomotion scenarios and environments for the quadruped. This contrasts with common evaluation protocols in the physics informed literature that train on partial trajectories and evaluate on disjoint temporal segments of the same underlying trajectories~\cite{tapley2024symplecticneuralnetworksbased,gruber2025efficientlyparameterizedneuralmetriplectic}.

We compare \textsc{CaLiSym} against both black-box and physics-informed baselines, including an MLP, a Transformer, teacher-forcing-trained RWM-TF~\cite{li2025roboticworldmodelneural}, a Dissipative Hamiltonian Neural Network (DHNN)~\cite{sosanya2022dhnn}, and Dissipative SymODEN (D-SymODEN)~\cite{zhong2020dissSymODEN}. The Transformer and RWM-TF operate on a 32-step context window and therefore receive access to multiple past observations, whereas all other methods, including \textsc{CaLiSym}, predict using only the current state and port variables.

Unless otherwise stated, \textsc{CaLiSym} is instantiated using either \textsc{GRB-SympNet} or \textsc{GR-SympNet} depending on the dimensionality of the lifted state space. All results are obtained under autoregressive rollout, the regime most relevant to planning and model-predictive control. We report out-of-distribution autoregressive mean-squared error (OOD AR MSE), parameter count, and floating-point operations per prediction (FLOPs), capturing long-horizon accuracy alongside computational efficiency. FLOPs are measured at the operator level over a single steady-state prediction and reflect what each architecture actually executes inside a control loop. The Transformer re-encodes its entire 32-step context window whenever it is queried, as attention carries no persistent memory, whereas RWM-TF merely advances its recurrence by one update. DHNN and D-SymODEN additionally pay for differentiating their learned Hamiltonians at inference time. For \textsc{CaLiSym} the total spans the full lift, symplectic evolution, and projection. 

In addition to predictive accuracy and computational efficiency, we verify empirically the exact preservation of the lifted symplectic structure. We therefore report the normalized symplecticity residual $\epsilon_{\mathbf J_{\boldsymbol \Omega}}$, which remains at machine precision across all of our models (Appendix~\ref{app:symplecticity_conservation}), confirming that the geometric constraint is satisfied by construction rather than approximately through optimization.

\subsection{Controlled Dissipative Double Pendulum}
\label{subsec:experiments_pendulum}
We first evaluate \textsc{CaLiSym} on a controlled dissipative double pendulum, a benchmark that isolates the central challenge addressed by this work, i.e., learning non-conservative, complex dynamics using a symplectic lifted representation. The system consists of two coupled rotational degrees of freedom subject to gravity, elasticity, viscous damping, and external torque inputs as shown in~Table~\ref{tab:pendulum_physics}. Although the underlying mechanical dynamics originate from a Hamiltonian system, damping and control inputs continuously inject and dissipate energy, making the observed dynamics fundamentally non-conservative.

\begin{table}[ht]
\vspace{-2mm}
\caption{Double pendulum physical parameters.}
\label{tab:pendulum_physics}
\centering
\small
\begin{tabular}{l l}
\toprule
Parameter & Value \\
\midrule
($m_1$, $m_2$) [$kg$] & (2.1, 1.3) \\
($l_1$, $l_2)$ [$m$] & (0.4, 1.3) \\
($I_1$, $I_2$) [$kg\cdot m^2$] & (1.1, 0.7) \\
($k_1$, $k_2$) [$N\,m/rad$] & (0.03, 0.06) \\
($b_1$, $b_2$) [$N \cdot m \cdot s/rad$] & (0.3, 0.6) \\
$g$ [$m/s^2$] & 9.81 \\
\bottomrule
\end{tabular}
\end{table}

The measured state is
\begin{equation}
\boldsymbol{x}=(\boldsymbol{q},\boldsymbol{p}) \in \mathbb{R}^{4},
\end{equation}
where $\boldsymbol{q}=(q_1,q_2)$ denotes the joint angles and $\boldsymbol{p}=(p_1,p_2)$ the corresponding generalized momenta. The physical state and control torques are embedded into the lifted section
\begin{equation}
\boldsymbol{Q}=(\boldsymbol{q},\boldsymbol{0},\boldsymbol{0}) \in \mathbb R^6,
\qquad
\boldsymbol{P}=(\boldsymbol{0},\boldsymbol{p},\boldsymbol{\tau}) \in \mathbb R^6,
\end{equation}
and evolved using \textsc{GRB-SympNet} within the lifted symplectic phase space.
Figure~\ref{fig:pendulum} shows representative autoregressive rollouts together with a quantitative comparison against all baselines. \textsc{CaLiSym} accurately tracks both joint angles and generalized momenta over long prediction horizons despite the presence of actuation and dissipation. Quantitatively, \textsc{CaLiSym} achieves an OOD AR MSE of $2.341$, compared to $7.688$ for the strongest competing baseline, corresponding to a $69.5\%$ reduction in rollout error. This improvement is obtained using only approximately $3.7$K parameters, whereas competing methods require roughly $10$K parameters. The results suggest that the lifted symplectic representation successfully resolves the mismatch between non-conservative physical dynamics and conventional symplectic learning, while the local spline approximation of \textsc{GRB-SympNet} provides sufficient expressivity to capture the resulting dynamics with a compact model.

\begin{figure*}[ht]
  \centering
  \includegraphics[width=0.192\textwidth, trim=5 0 8 0, clip]{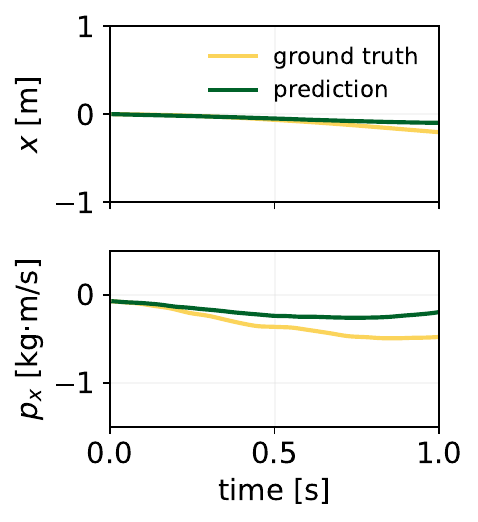}\hfill
  \includegraphics[width=0.192\textwidth, trim=5 0 8 0, clip]{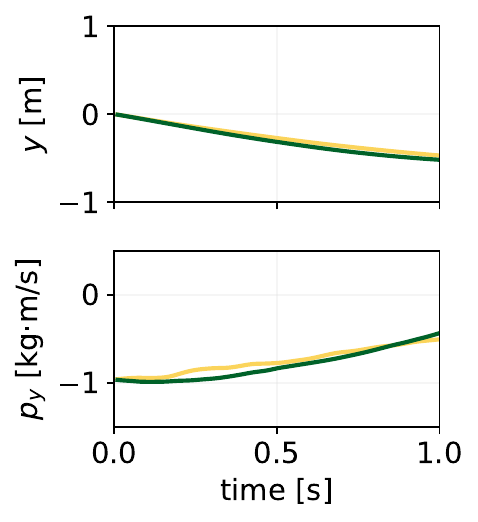}\hfill
  \includegraphics[width=0.192\textwidth, trim=5 0 8 0, clip]{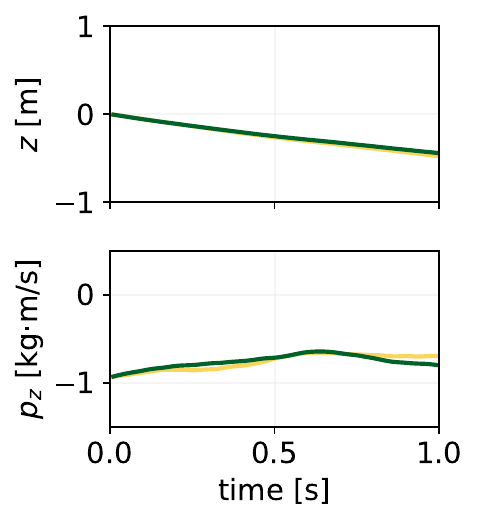}\hfill
  \includegraphics[width=0.424\textwidth, trim=-10 0 20 40, clip]{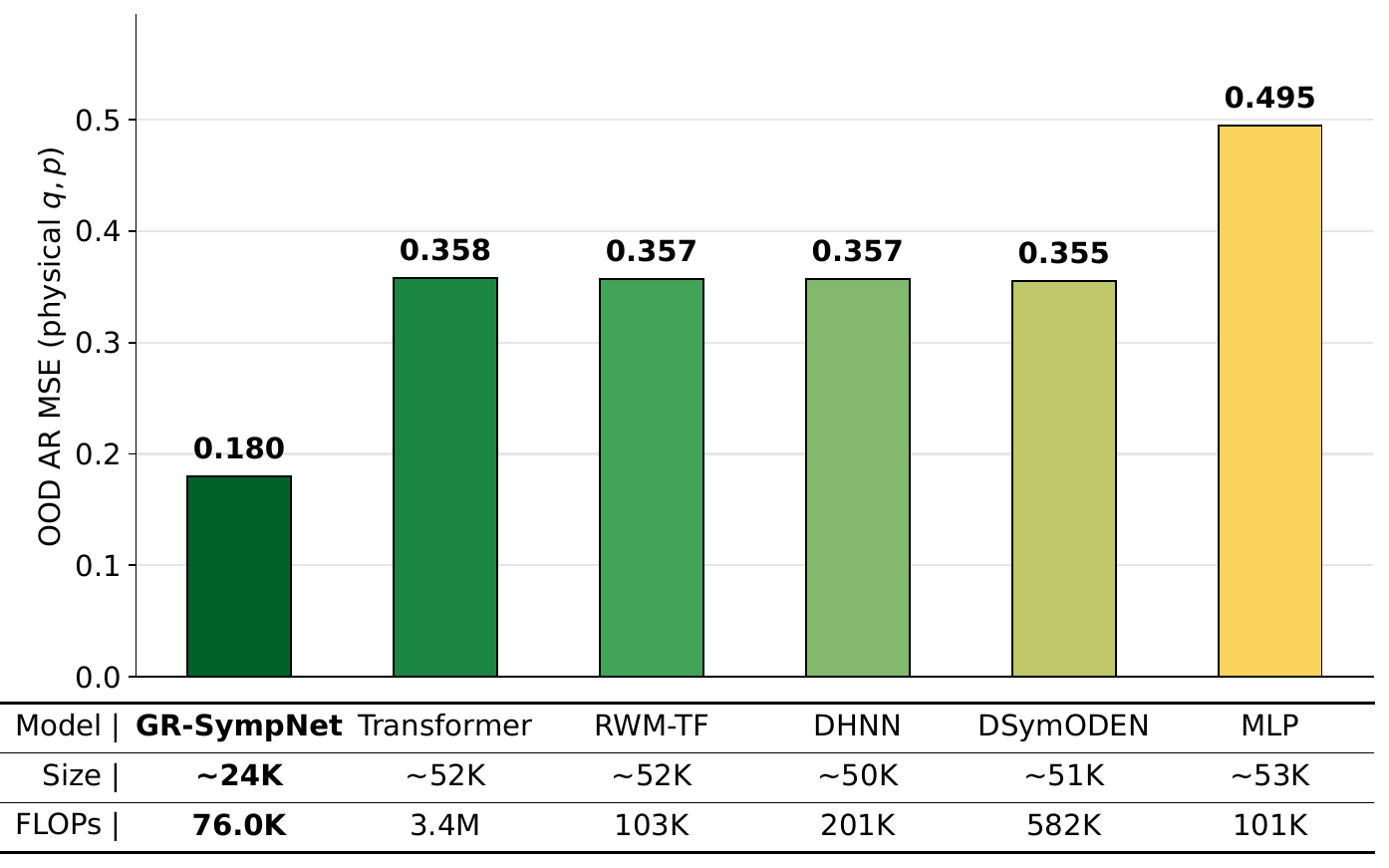}\hfill
  \vspace{-2mm}
  \caption{\textbf{Quadrotor dynamics.} An underactuated aerial forced system with rotor thrust, aerodynamic and other unmodeled effects. (\textbf{Left}) An autoregressive 100-timestep out-of-distribution rollout of the body position $(x,y,z)$ and corresponding momenta. (\textbf{Right}) The bar plot reports out-of-distribution autoregressive MSE and model size. \textsc{CaLiSym} achieves the lowest OOD rollout error while remaining parameter-efficient, showing that the same lift transfers without modification to forced aerial dynamics. It is also the cheapest to evaluate, needing $76.0K$ FLOPs per step, $24.8\%$ less than the nearest baseline.}
  \label{fig:quadrotor}
  \vspace{-5mm}
\end{figure*}

\vspace{-2mm}
\subsection{Quadrotor Dynamics}
\label{subsec:results_quadrotor} 
The quadrotor experiment evaluates whether the proposed structured lift remains effective for underactuated aerial systems with strongly control-dependent dynamics. Unlike the quadruped, where energy and momentum are exchanged through both actuation and intermittent contact, the quadrotor interacts with its environment primarily through rotor thrust and aerodynamic effects. Rotor inputs continuously inject energy into the system, while aerodynamic drag and other unmodeled effects introduce dissipation that is difficult to capture through purely conservative formulations. The used real-world dataset~\cite{mohajerin18quad, moharejin19quadrnn} comprises 54 flights of a quadrotor with parameters as shown in Table~\ref{tab:quadrotor_params}.
\begin{table}[ht]
  \centering
  \caption{Quadrotor (AscTec Pelican) physical parameters.}
  \label{tab:quadrotor_params}
  \begin{tabular}{llll}
    \toprule
    Parameter & Value \\
    \midrule
    $m$  [$kg$]                       & $1.6$  \\
    $(I_{xx}, I_{yy}, I_{zz})$ [$kg\,m^2$] & $(0.002, 0.002, 0.001)$  \\
    $k_f$                      & $9.9865 \times 10^{-6}$  \\
    $k_m$                      & $1.5978 \times 10^{-7}$  \\
    $l$ [$m$]                        & $0.21$ \\
    \bottomrule
  \end{tabular}
\end{table}

The physical state consists of the vehicle pose, orientation, and generalized momenta. The available ports correspond to the commanded rotor thrusts and body torques. \textsc{CaLiSym} embeds the physical state together with the control ports into the lifted section,
\begin{equation}
    \mathbf Q = (\mathbf q,\mathbf0,\mathbf0)\in \mathbb R^{18},\quad \mathbf P = (\mathbf 0,\mathbf p,\boldsymbol\tau)\in \mathbb R^{18},
\end{equation}
where $\boldsymbol \tau$ denotes the control inputs. The lifted dynamics are evolved using an exactly symplectic \textsc{GR-SympNet} predictor before being projected back onto the physical state space through the gauge-fixed rollout procedure described in Section~\ref{sec:methods}.

This experiment is particularly challenging because quadrotor dynamics are highly sensitive to control inputs and rapidly accumulate prediction errors during autoregressive rollout. Small inaccuracies in attitude prediction can quickly propagate into translational motion, resulting in substantial trajectory drift. Consequently, the task provides a stringent test of whether the proposed lifted representation can maintain stable long-horizon predictions despite the strongly forced and dissipative nature of the underlying dynamics.

Figure~\ref{fig:quadrotor}  shows representative autoregressive rollouts together with the corresponding OOD comparison. \textsc{CaLiSym} achieves the lowest OOD rollout error among the evaluated methods while remaining parameter-efficient. More importantly, the results demonstrate that the same lift formulation used for the double pendulum transfers directly to aerial dynamics without modification, supporting the claim that the structured lift is a general mechanism for extending symplectic learning beyond conservative systems.

\vspace{-4mm}
\subsection{Quadruped Dynamics} \label{subsec:experiments_quadruped} 

The quadruped experiment evaluates whether the proposed lifted symplectic representation remains effective in the high-dimensional, contact-rich regime characteristic of real robotic systems. The physical state comprises floating-base and joint coordinates together with their corresponding generalized momenta~\cite{fan2024review}. We evaluate \textsc{CaLiSym} on the \textsc{GrandTour} locomotion dataset~\cite{frey2026grandtour}, which contains extensive real-world trajectories of a quadruped robot performing diverse locomotion behaviors across varying commands, terrains, and environmental conditions. This benchmark poses a particularly challenging dynamics-learning problem due to the combination of floating-base motion, intermittent contacts, impacts, actuator dynamics, contact slippage, and rapidly changing contact modes, all of which can amplify small modeling errors during long-horizon autoregressive rollout.

\textsc{CaLiSym} represents these effects by embedding control torques and contact forces into the lifted section
\begin{equation}
\boldsymbol{Q}=(\boldsymbol{q},\boldsymbol{0},\boldsymbol{0},\boldsymbol{f}_c)\in\mathbb{R}^{66},
\qquad
\boldsymbol{P}=(\boldsymbol{0},\boldsymbol{p},\bar{\boldsymbol{\tau}},\boldsymbol{0})\in\mathbb{R}^{66},
\end{equation}
where $\bar{\boldsymbol{\tau}}=(\boldsymbol{0}_6,\boldsymbol{\tau})$ augments the joint torques with zeros corresponding to the unactuated floating-base coordinates. The lifted dynamics are modeled using \textsc{GR-SympNet}, whose generalized-ridge parameterization scales efficiently to the resulting high-dimensional state space.

\begin{figure*}[ht]
  \centering
  \includegraphics[width=0.183\textwidth, trim=5 0 8 0, clip]{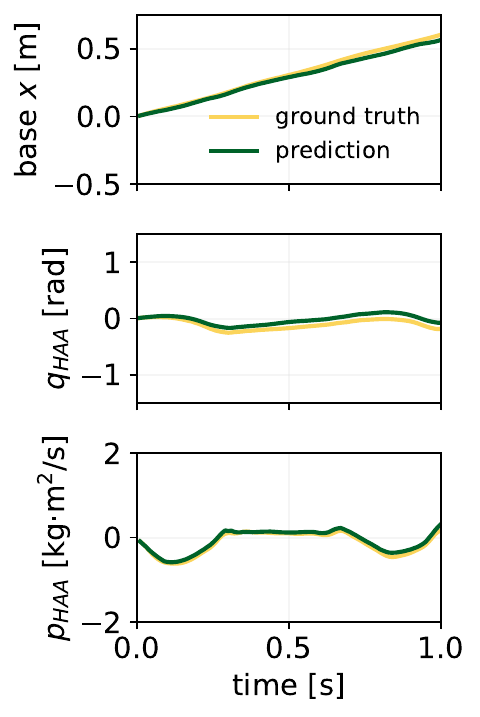}\hfill
  \includegraphics[width=0.183\textwidth, trim=5 0 8 0, clip]{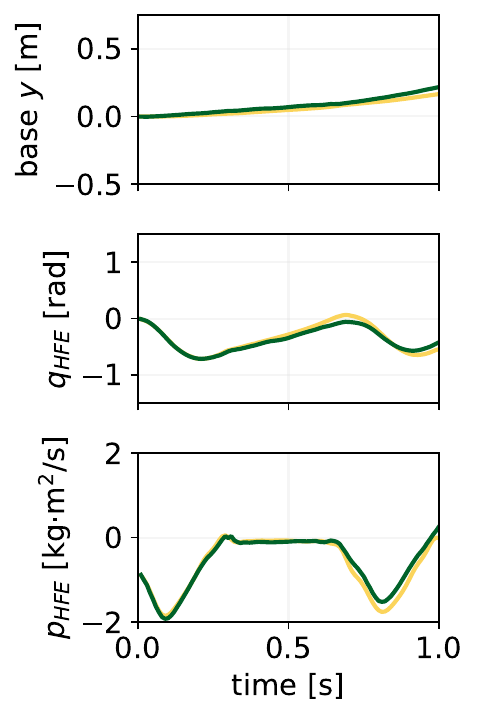}\hfill
  \includegraphics[width=0.183\textwidth, trim=5 0 8 0, clip]{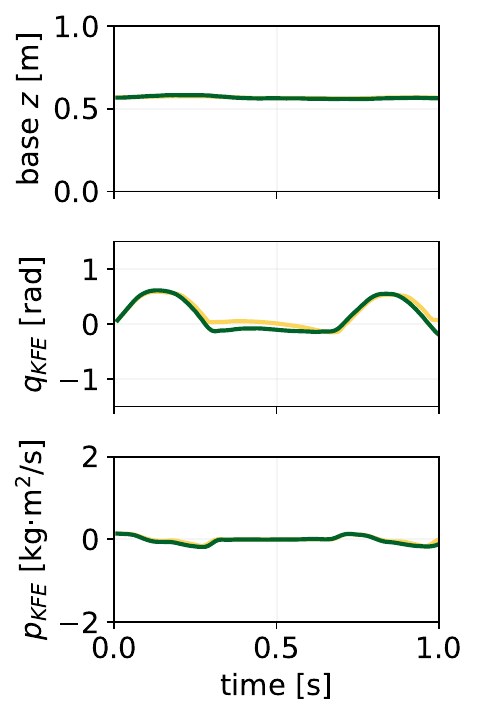}\hfill
  \includegraphics[width=0.45\textwidth, trim=-10 0 20 0, clip]{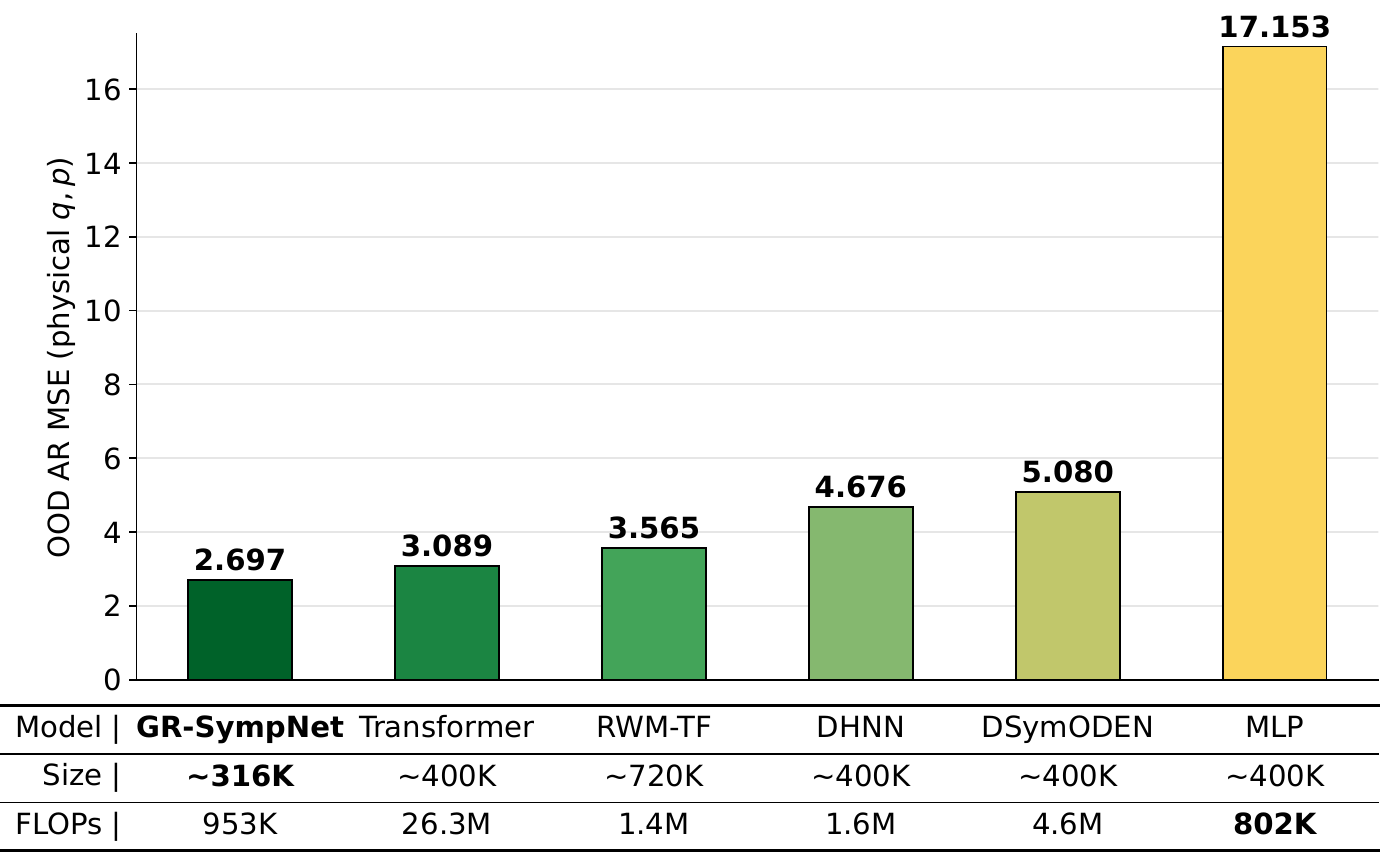}\hfill
  \vspace{-2mm}
    \caption{\textbf{Quadruped dynamics.} A high-dimensional, contact-constrained floating-base system actuated through joint torques and ground contact. (\textbf{Left}) An autoregressive out-of-distribution rollout of the floating-base position and representative right-leg coordinates and momenta, where $q_{\mathrm{HAA}}$, $q_{\mathrm{HFE}}$, $q_{\mathrm{KFE}}$ denote hip abduction/adduction, hip flexion/extension, and knee flexion/extension, with conjugate momenta $p_{\mathrm{HAA}}$, $p_{\mathrm{HFE}}$, $p_{\mathrm{KFE}}$. \textsc{CaLiSym} embeds joint torques and contact forces into the lift, keeping the projected dynamics contact-constrained and non-conservative. (\textbf{Right}) The bar plot reports out-of-distribution autoregressive MSE and model size. \textsc{CaLiSym} with \textsc{GR-SympNet} achieves the lowest OOD error ($2.697$ vs.\ $3.089$ for the strongest baseline, a $\sim$12.7\% reduction) with $\sim$316K parameters, fewer than the main baselines (400K--720K). Evaluating one step costs 953K FLOPs, far less than any of the sequence models. The MLP is slightly cheaper because it acts on the raw 60-dimensional state, whereas \textsc{CaLiSym} works in the 132-dimensional lifted space that carries the torque and contact channels. That overhead amounts to less than 19\% and buys a $6.4\times$ lower rollout error.}
  \label{fig:quadruped}
  \vspace{-5mm}
\end{figure*}

Figure~\ref{fig:quadruped} presents representative autoregressive rollouts together with an OOD comparison against all baselines. The rollout trajectories demonstrate that \textsc{CaLiSym} accurately captures both the dominant floating-base motion and the evolution of representative leg joints despite the heterogeneous scales of base coordinates, joint angles, contact forces, and generalized momenta. Quantitatively, \textsc{CaLiSym} achieves the lowest OOD autoregressive error among all evaluated methods, obtaining an OOD AR MSE of $2.697$ compared to $3.089$ for the Transformer, $3.565$ for RWM-TF, $4.676$ for DHNN, and $5.080$ for D-SymODEN. This corresponds to a $12.7\%$ reduction relative to the strongest baseline and a $46.9\%$ reduction relative to D-SymODEN.

Importantly, these improvements are achieved with smaller network sizes compared to the baselines. \textsc{CaLiSym} uses approximately $316$K parameters, compared to roughly $400$K parameters for the Transformer, DHNN, and D-SymODEN, and approximately $720$K parameters for RWM-TF. The results therefore suggest that the gains arise from the lifted symplectic representation and the associated symplectic inductive bias rather than large model size. Moreover, they demonstrate that the generalized-ridge architecture of \textsc{GR-SympNet} can scale the proposed framework to realistic contact-rich robotic systems without resorting to large autoregressive sequence models with large context windows.

\vspace{-4mm}
\subsection{Summary}
\label{subsec:experiments_summary}

Across the double pendulum, quadrotor, and quadruped benchmarks, the results consistently support the central hypotheses of this work. First, learning dynamics in a structured lifted symplectic space improves long-horizon prediction for non-conservative systems. Despite actuation, dissipation, and contact interactions, \textsc{CaLiSym} achieves the lowest OOD autoregressive rollout error across all three systems.

Second, these improvements come with strictly smaller models. \textsc{CaLiSym} uses roughly $3.7K$ parameters against $10K$ for every baseline on the double pendulum, $24K$ against $\sim50K$ on the quadrotor, and $316K$ against $400K$ to $720K$ on the quadruped, ruling out capacity as the main source of the gains. The comparison is if anything conservative, since the Transformer and RWM-TF additionally condition on a $32$-step context window while \textsc{CaLiSym} predicts from the current state and ports alone (fully Markovian rollout). That both \textsc{GRB-SympNet} and \textsc{GR-SympNet} outperform these larger baselines at every scale indicates the gains arise from the geometric inductive bias of the lifted representation, not from capacity or access to history.

Third, the same lift-evolve-project-re-embed procedure transfers across systems with substantially different dimensionality and interaction mechanisms, from a controlled dissipative pendulum to an aerial quadrotor and an underactuated quadruped. Empirically, parameter count scales quadratically with the lifted phase-space dimension ($d$) across the three systems, consistent with $\Theta(md)$ ridge matrices whose width $m$ is chosen proportional to $d$. Fixing the ridge dimension instead would recover linear scaling, at the cost of per-layer expressivity.

Taken together, these results demonstrate that symplectic dynamics learning is not confined to conservative Hamiltonian systems. The lift is what makes this possible. Evolution in the lifted phase space remains symplectic by construction, so the geometric prior responsible for stability is preserved regardless of what happens physically, and control, dissipation, and constraint enter only through the projection back to the physical state. By enforcing symplectic structure in this lifted space while leaving the induced physical dynamics free to be controlled, dissipative, or constrained, \textsc{CaLiSym} unites the stability of geometric learning with the flexibility that real-world robotics demands.

\vspace{-2mm}
\section{Conclusion}
\label{sec:conclusion}
We introduced \textsc{CaLiSym}, a framework for learning symplectic representations of robotic systems whose observed dynamics are controlled, dissipative, and constrained. The central idea is to enforce symplectic structure in a structured lifted canonical phase space rather than directly on the measured physical state, allowing the induced physical dynamics to remain non-conservative.

The proposed framework separates geometric structure from physical energy exchange. Dynamics are learned as exact symplectic maps in the lifted space, while projection and re-embedding ensure consistency with the observed state, controls, and constraints. Because the ports enter as explicit canonical variables, this separation also exposes the energy exchange between the robot and its environment as a first-class quantity rather than an implicit byproduct of the fit, opening a path from learned dynamics toward passivity-based and certificate-based control synthesis~\cite{vanderschaft2017l2gain, brunke2022safe}. The model predicts from the current state and ports alone, which is the information a receding-horizon solver already holds at each step of its internal rollout, so there is no observation history to maintain and no context window to replay at every iteration. This construction retains the computational advantages of explicit SympNet-style predictors while extending their applicability beyond closed Hamiltonian systems.

Across controlled dissipative and contact-constrained robotic systems, \textsc{CaLiSym} consistently improves out-of-distribution autoregressive prediction while remaining parameter-efficient. The predictor achieves this at parameter counts and inference costs compatible with control rates on embedded hardware, making the underlying guarantees practically realizable rather than a purely theoretical consideration. These results demonstrate that symplectic learning can be extended beyond conservative systems through a structured canonical lift, providing a practical geometric inductive bias for long-horizon robot dynamics prediction.

Future work will investigate richer port representations, uncertainty-aware lifted dynamics, and integration with model-predictive control, trajectory optimization, and planning pipelines. More generally, this work suggests that physically meaningful geometric structure can be recovered and exploited even when it is not apparent in the measured dynamics, opening new opportunities for structure-preserving learning in complex robotic systems.
\vspace{-2mm}
\appendices

\section{Autoregressive Rollout and Multi-Horizon Evaluation}
\label{app:multi-horizon}
The main experiments evaluate \textsc{CaLiSym} autoregressively, querying the model on its own predictions, so errors compound along the horizon in a way a one-step metric cannot reveal. To characterize this, we roll out the same trained lifted predictor on the controlled dissipative double pendulum over horizons from $5$ to $800$ steps at a fixed $\Delta t = 0.01,\mathrm{s}$ ($0.05$ to $8.00,\mathrm{s}$), reporting the in-distribution (ID) and out-of-distribution (OOD) total mean-squared error over the full lifted state, rather than the physical $(q,p)$ error alone~(Table~\ref{tab:multihorizon}).

The ID error stays small across the full sweep, rising from $\sim\!10^{-5}$ at $5$ steps to $\sim\!10^{-2}$ at $800$ steps with no runaway growth, so the lifted map remains numerically stable over an $8\,\mathrm{s}$ horizon despite the driven, dissipative dynamics. The OOD error grows steeply, as expected. Once a trajectory leaves the training distribution, each step is conditioned on an increasingly model generated state, compounding distribution shift error. The two effects are distinct. One is bounded numerical stability from the symplectic structure, and the other is error growth from distribution shift, rather than a single rollout failure.
\vspace{-3mm}
\begin{table}[ht]
\centering
\caption{Total ID and OOD multi-horizon rollout error for lifted state.}
\label{tab:multihorizon}
\sisetup{table-format=1.3e1}
\begin{tabular}{ccSc}
\toprule
Timesteps & $\Delta \mathrm T$ [s] & {ID MSE} & {OOD MSE} \\
\midrule
5   & 0.05 & 1.130e-5 & 0.070 \\
10  & 0.10 & 2.616e-5 & 0.258  \\
20  & 0.20 & 7.027e-5 & 0.718   \\
50  & 0.50 & 2.478e-4 & 1.771   \\
100 & 1.00 & 2.591e-4 & 4.892   \\
200 & 2.00 & 6.396e-4 & 14.341  \\
400 & 4.00 & 7.922e-4 & 39.553  \\
600 & 6.00 & 0.033 & 69.460     \\
800 & 8.00 & 0.033 & 101.397    \\
\bottomrule
\end{tabular}
\end{table}

\section{Symplecticity Conservation}
\label{app:symplecticity_conservation}
The defining property of the learned lifted predictor is exact symplecticity in the lifted canonical phase space. For a differentiable map $\Phi_\theta^{\Delta t}$ on lifted coordinates $\boldsymbol{Z}$, symplecticity means
\begin{equation}
    D\Phi_\theta^{\Delta t}(\boldsymbol{Z})^\top \mathbf{J}_{\Omega}
    D\Phi_\theta^{\Delta t}(\boldsymbol{Z}) = \mathbf{J}_{\Omega},
\end{equation}
where $\mathbf{J}_{\Omega}$ is the canonical symplectic matrix of the lifted
space. \textsc{GR-SympNet} and \textsc{GRB-SympNet} satisfy this by construction, since each layer is the exact flow of a Hamiltonian system and the composition of symplectic maps remains symplectic.

We still verify the property numerically for the trained models (Table~\ref{tab:symplecticity_check}), for two reasons. First, it confirms that the trained architecture realizes the intended geometry rather than approximating it. Second, it separates a hard architectural constraint from a soft-regularization penalty, since a structurally symplectic map sits at numerical precision regardless of the training loss, whereas a penalized one would track it. For each model we compute the normalized residual,
\begin{equation}
    \epsilon_{\mathbf{J}_{\Omega}} =
    \frac{ \left\| D\Phi_\theta^{\Delta t}(\mathbf{Z})^\top \mathbf{J}_{\Omega}
    D\Phi_\theta^{\Delta t}(\mathbf{Z}) - \mathbf{J}_{\Omega} \right\|_F }
    { \|\mathbf{J}_{\Omega}\|_F }.
\label{eq:symplecticity_residual}
\end{equation}
The Jacobian is taken with respect to the lifted state $\mathbf Z$, not the projected physical state. This distinction is central to \textsc{CaLiSym}, as the physical predictor induced by lifting, projecting, and embedding back is not required to be symplectic on the measured state. Symplecticity is enforced only on the lifted map.

\begin{table}[ht]
\vspace{-4mm}
\centering
\caption{Symplecticity residuals are at numerical precision.}
\label{tab:symplecticity_check}
\renewcommand{\arraystretch}{1.15}
\begin{tabular}{@{}l l c c c@{}}
\toprule
\textbf{System}
&
\textbf{Architecture}
&
\textbf{Dim.}
&
\boldmath$\epsilon_{\mathbf{J}_{\Omega}}$\textbf{ mean}
&
\boldmath$\epsilon_{\mathbf{J}_{\Omega}}$\textbf{ max}
\\
\midrule
Double pendulum
&
\textsc{GRBSympNet}
&
$12$
&
$3\times 10^{-16}$
&
$4\times 10^{-15}$
\\
Quadrotor
&
\textsc{GRSympNet}
&
$36$
&
$3\times 10^{-15}$
&
$8\times 10^{-15}$
\\
Quadruped
&
\textsc{GRSympNet}
&
$132$
&
$5\times 10^{-16}$
&
$8\times 10^{-16}$
\\
\bottomrule
\end{tabular}
\vspace{-5mm}
\end{table}

\section{Universal Approximation of \textsc{GRB-SympNet}}
\label{app:grb_sympnet_uat}

This appendix states the approximation property of the \textsc{GRB-SympNet}
predictor used for compact low-dimensional lifted dynamics. The result is
local in a canonical coordinate chart of the lifted phase space. This is
the relevant setting for \textsc{CaLiSym}, where symplecticity is imposed on the
lifted map
$$
\Phi_\theta^{\Delta t}:\mathcal Z\to\mathcal Z,
$$
while the induced physical predictor
$$
\Pi_X\circ\Phi_\theta^{\Delta t}
\circ\sigma_{\mathbf e_k}:X\to X
$$
need not be symplectic on the measured physical state space
$X:=T^*\mathcal Q$.

Let
$\mathbf Z=(\mathbf Q,\mathbf P)\in\mathcal Z\simeq\mathbb R^{2d}$,
with canonical symplectic form
$\Omega=\sum_i dQ^i\wedge dP_i$ and canonical symplectic matrix
$\mathbf J_\Omega$. For a compact set
$\mathcal K\subset\mathcal Z$, write
$\|\cdot\|_{C^s(\mathcal K)}$ for the maximum over derivatives up to
order $s$. To avoid conflict with the auxiliary lifted coordinate
$\mathbf r$, we denote the B-spline degree in this appendix by
$r_{\rm spl}$.

A \textsc{GRB-SympNet} layer is the exact generalized-ridge Hamiltonian flow
$$
\begin{aligned}
\boldsymbol\xi_\ell
&=
\mathbf A_\ell\mathbf P
+
\mathbf B_\ell\mathbf Q,\\
\mathbf Q'
&=
\mathbf Q
+
h_\ell\mathbf A_\ell^\top
\nabla K_\ell(\boldsymbol\xi_\ell),\\
\mathbf P'
&=
\mathbf P
-
h_\ell\mathbf B_\ell^\top
\nabla K_\ell(\boldsymbol\xi_\ell),\\
\mathbf A_\ell\mathbf B_\ell^\top
&=
\mathbf B_\ell\mathbf A_\ell^\top .
\end{aligned}
$$
Equivalently,
$\varphi_\ell(\mathbf Q,\mathbf P)=(\mathbf Q',\mathbf P')$.
The scalar ridge Hamiltonian is parameterized as
$$
\begin{aligned}
K_\ell(\boldsymbol\xi)
&=
K_{\ell,\mathrm{spline}}(\boldsymbol\xi)
+
K_{\ell,\mathrm{smooth}}(\boldsymbol\xi),\\
K_{\ell,\mathrm{spline}}(\boldsymbol\xi)
&=
\sum_{\alpha\in\mathcal A}
c_{\ell,\alpha}\,
\mathcal B_\alpha(\boldsymbol\xi),\\
\mathcal B_\alpha(\boldsymbol\xi)
&=
\prod_{j=1}^{m}
B_{\alpha_j}^{(r_{\rm spl})}(\xi_j).
\end{aligned}
$$
The smooth tail is
$$
\begin{aligned}
K_{\ell,\mathrm{smooth}}(\boldsymbol\xi)
&=
\sum_{j=1}^{N_s}
a_{\ell j}\,\phi(\rho_{\ell j}),\\
\rho_{\ell j}
&=
\mathbf w_{\ell j}^\top\boldsymbol\xi
+
b_{\ell j}.
\end{aligned}
$$
The smooth tail is useful in the experimental model class, but it is not
needed for the density argument. Setting
$K_{\ell,\mathrm{smooth}}\equiv0$ gives a subclass of the full
\textsc{GRB-SympNet} family.

\noindent\textbf{Theorem.}
\emph{Lifted universal approximation by full-ridge \textsc{GRB-SympNet}s.}
Let $\mathcal U\subset\mathcal Z\simeq\mathbb R^{2d}$ be open, let
$\mathcal K\Subset\mathcal U$ be compact, and let
$\Psi:\mathcal U\to\mathcal Z$ be a $C^{s+1}$ symplectic map, meaning
$$
\begin{aligned}
D\Psi(\mathbf Z)^\top
\mathbf J_\Omega
D\Psi(\mathbf Z)
&=
\mathbf J_\Omega,\\
\mathbf Z&\in\mathcal U .
\end{aligned}
$$
Assume that the generalized-ridge dimension may be chosen as $m=d$, and
assume that the B-spline spaces used for
$K_{\ell,\mathrm{spline}}$ are $C^{s+1}$-dense on compact subsets of
$\mathbb R^d$. For simple knots, it is sufficient to take
$r_{\rm spl}\ge s+2$. Then, for every $\varepsilon>0$, there exists a
finite-depth \textsc{GRB-SympNet}
$$
\Phi_\theta^{\Delta t}
=
\varphi_L\circ\cdots\circ\varphi_1
$$
such that
$$
\|\Phi_\theta^{\Delta t}-\Psi\|_{C^s(\mathcal K)}
<
\varepsilon
$$
and
$$
\begin{aligned}
&D\Phi_\theta^{\Delta t}(\mathbf Z)^\top
\mathbf J_\Omega
D\Phi_\theta^{\Delta t}(\mathbf Z)\\
&\hspace{2.5em}
=
\mathbf J_\Omega .
\end{aligned}
$$
Thus \textsc{GRB-SympNet}s are dense, on compact subsets of the lifted canonical
phase space, in the class of smooth symplectic maps, while remaining
exactly symplectic for every parameter value.

\paragraph{Proof.}
First, each \textsc{GRB-SympNet} layer is exactly symplectic. The layer is the
time-$h_\ell$ flow of the scalar Hamiltonian
$$
\begin{aligned}
H_\ell(\mathbf Q,\mathbf P)
&=
K_\ell(
\mathbf A_\ell\mathbf P
+
\mathbf B_\ell\mathbf Q)\\
&=
K_\ell(\boldsymbol\xi_\ell).
\end{aligned}
$$
Hamilton's equations give
$$
\begin{aligned}
\dot{\mathbf Q}
&=
\mathbf A_\ell^\top
\nabla K_\ell(\boldsymbol\xi_\ell),\\
\dot{\mathbf P}
&=
-
\mathbf B_\ell^\top
\nabla K_\ell(\boldsymbol\xi_\ell).
\end{aligned}
$$
Hence
$$
\begin{aligned}
\dot{\boldsymbol\xi}_\ell
&=
\mathbf A_\ell\dot{\mathbf P}
+
\mathbf B_\ell\dot{\mathbf Q}\\
&=
\left(
-\mathbf A_\ell\mathbf B_\ell^\top
+
\mathbf B_\ell\mathbf A_\ell^\top
\right)
\nabla K_\ell(\boldsymbol\xi_\ell)\\
&=
\mathbf 0 .
\end{aligned}
$$
Thus $\boldsymbol\xi_\ell$ is constant along the layer flow. The
Hamiltonian flow therefore integrates exactly to the explicit GRB layer
above. Since exact Hamiltonian flows preserve $\Omega$, each
$\varphi_\ell$ is symplectic. The composition
$\Phi_\theta^{\Delta t}$ is symplectic because symplectic maps are closed
under composition.

It remains to prove density. In the full-ridge case $m=d$, the GRB layer
contains the two standard gradient shears. For
$(\mathbf A,\mathbf B)=(\mathbf I_d,\mathbf 0)$,
$$
S_K^Q(\mathbf Q,\mathbf P)
=
(\mathbf Q+\nabla K(\mathbf P),\mathbf P).
$$
For $(\mathbf A,\mathbf B)=(\mathbf 0,\mathbf I_d)$,
$$
S_K^P(\mathbf Q,\mathbf P)
=
(\mathbf Q,\mathbf P-\nabla K(\mathbf Q)).
$$
Let $K_0(\mathbf u)=\frac12\|\mathbf u\|^2$. The canonical symplectic
rotation
$$
R(\mathbf Q,\mathbf P)
=
(\mathbf P,-\mathbf Q)
$$
factors as
$$
R
=
S_{K_0}^P
\circ
S_{K_0}^Q
\circ
S_{K_0}^P .
$$
Indeed, the three shears map
$(\mathbf Q,\mathbf P)$ to
$(\mathbf Q,\mathbf P-\mathbf Q)$, then to
$(\mathbf P,\mathbf P-\mathbf Q)$, and finally to
$(\mathbf P,-\mathbf Q)$.

For a scalar potential $V:\mathbb R^d\to\mathbb R$ and a constant vector
$\boldsymbol\eta\in\mathbb R^d$, define the H\'enon-like symplectic map
$$
H_{V,\boldsymbol\eta}(\mathbf Q,\mathbf P)
=
(\mathbf P+\boldsymbol\eta,
-\mathbf Q+\nabla V(\mathbf P)).
$$
It is represented using GRB shears by
$$
\begin{aligned}
H_{V,\boldsymbol\eta}
&=
T_{\boldsymbol\eta}^Q
\circ
R
\circ
S_{-V}^Q,\\
T_{\boldsymbol\eta}^Q(\mathbf Q,\mathbf P)
&=
(\mathbf Q+\boldsymbol\eta,\mathbf P).
\end{aligned}
$$
The translation $T_{\boldsymbol\eta}^Q$ is generated by the linear
potential $K(\mathbf P)=\boldsymbol\eta^\top\mathbf P$, while the rotation
is generated by quadratic shear potentials. These linear and quadratic
potentials are contained in, or can be approximated arbitrarily well by,
the spline Hamiltonian class on compact sets.

By the H\'enon approximation theorem for symplectic maps, and by the same
approximation route underlying SympNet universality
\cite{jin2020sympnets}, there exist smooth scalar
potentials $V_1,\ldots,V_N$ and constant vectors
$\boldsymbol\eta_1,\ldots,\boldsymbol\eta_N$ such that
$$
\begin{aligned}
\mathcal H
&= H_{V_N,\boldsymbol\eta_N}
\circ\cdots\circ
H_{V_1,\boldsymbol\eta_1},\\
&\|\Psi-\mathcal H\|_{C^s(\mathcal K)}
<
\frac{\varepsilon}{2}.
\end{aligned}
$$
Let $\mathcal K_j$ be compact neighborhoods of the image sets encountered
by this finite composition, and let $\pi_P$ denote projection onto the
$\mathbf P$-coordinates. Since the spline Hamiltonian class is
$C^{s+1}$-dense, choose spline potentials $K_{j,G}$ such that
$$
\|V_j-K_{j,G}\|_{C^{s+1}(\pi_P\mathcal K_j)}
<
\delta_j .
$$
Because $H_{V_j,\boldsymbol\eta_j}$ depends on $V_j$ only through
$\nabla V_j$, there is a compact-set constant $C_j$ with
$$
\begin{aligned}
&
\|H_{V_j,\boldsymbol\eta_j}
-
H_{K_{j,G},\boldsymbol\eta_j}
\|_{C^s(\mathcal K_j)} \le C_j\delta_j .
\end{aligned}
$$
Define the spline-composed H\'enon approximation
$$
\begin{aligned}
\widetilde{\mathcal H}_G
&=
H_{K_{N,G},\boldsymbol\eta_N}
\circ\cdots\circ H_{K_{1,G},\boldsymbol\eta_1}.
\end{aligned}
$$
Continuity of finite composition in the $C^s$ topology allows the
$\delta_j$ to be chosen so that
$$
\|\mathcal H-\widetilde{\mathcal H}_G\|_{C^s(\mathcal K)}
<
\frac{\varepsilon}{2}.
$$
The map $\widetilde{\mathcal H}_G$ is a \textsc{GRB-SympNet} because each
H\'enon-like factor is built from GRB shears with spline scalar
Hamiltonians. Hence the resulting network satisfies
$$
\|\Phi_\theta^{\Delta t}-\Psi\|_{C^s(\mathcal K)}
<
\varepsilon .
$$
Exact symplecticity follows from the layer construction. 

\paragraph{Spline-rate form.}
After fixing a H\'enon approximation
$$
\mathcal H
=
H_{V_N,\boldsymbol\eta_N}
\circ\cdots
\circ
H_{V_1,\boldsymbol\eta_1},
$$
the usual spline approximation rate gives a quantitative version of the
preceding argument. If each $V_j$ has sufficient smoothness on the
relevant compact set and the spline grid for $V_j$ has $G_j$ intervals in
each ridge coordinate, then
$$
\begin{aligned}
\|\Psi-\Phi_{\theta,G}^{\Delta t}\|_{C^s(\mathcal K)}
&\le
\varepsilon_{\rm Henon} + C_{\rm comp}\sum_{j=1}^N G_j^{-(r_{\rm spl}-s)} .
\end{aligned}
$$
Here
$\varepsilon_{\rm Henon}=\|\Psi-\mathcal H\|_{C^s(\mathcal K)}$ is the
finite H\'enon-composition error, and $C_{\rm comp}$ depends on the
compact image chain and derivative bounds of the composed maps. The loss
of one derivative is intrinsic: the symplectic layer uses
$\nabla K_\ell$, so $C^s$ approximation of the map requires
$C^{s+1}$ approximation of the scalar Hamiltonian.

\paragraph{Fixed-ridge interpretation.}
The theorem above is a full-ridge universal approximation result, since
it allows $m=d$. The practical \textsc{GRB-SympNet} regime may use $m\ll d$ for
parameter efficiency. In that case, the corresponding statement is
ridge-structured rather than fully universal. Suppose the generating
potentials in the H\'enon approximation admit additive generalized-ridge
structure,
$$
V_j(\mathbf P)
=
\sum_{\nu=1}^{R_j}
W_{j,\nu}(\mathbf C_{j,\nu}\mathbf P),
\qquad
\mathbf C_{j,\nu}\in\mathbb R^{m\times d}.
$$
Then
$$
\begin{aligned}
\nabla V_j(\mathbf P)
&=
\sum_{\nu=1}^{R_j}
\mathbf C_{j,\nu}^\top
\mathbf g_{j,\nu}(\mathbf P),\\
\mathbf g_{j,\nu}(\mathbf P)
&=
\nabla W_{j,\nu}
(\mathbf C_{j,\nu}\mathbf P).
\end{aligned}
$$
Each summand is exactly the update produced by a GRB layer with
$\mathbf A=\mathbf C_{j,\nu}$ and $\mathbf B=\mathbf 0$:
$$
\begin{aligned}
\mathbf Q'
&=
\mathbf Q
+
\mathbf C_{j,\nu}^\top
\nabla W_{j,\nu}
(\mathbf C_{j,\nu}\mathbf P),\\
\mathbf P'
&=
\mathbf P .
\end{aligned}
$$
Because these layers leave $\mathbf P$ fixed, their composition adds the
ridge-gradient contributions. The same argument applies to
$\mathbf P$-shears. Thus, for $m\ll d$, \textsc{GRB-SympNet} is universal for the
class of symplectic maps whose scalar generating potentials admit smooth
low-dimensional generalized-ridge decompositions. This is the regime
targeted by the low-dimensional double-pendulum experiment, where the
spline ridge parameterization remains compact.

\paragraph{Induced physical predictor}
The lifted result directly implies an approximation statement for the
\textsc{CaLiSym} physical predictor. Let $C_X\Subset X$ and $C_E\Subset E$ be
compact sets of physical states and ports. Suppose the one-step physical
transition
$$
f:X\times E\to X,
\qquad
f(\mathbf x,\mathbf e)=f_{\mathbf e}(\mathbf x),
$$
admits a smooth lifted symplectic realization on the data section. That
is, assume there exists a $C^{s+1}$ symplectic map $\Psi$ such that
$$
\begin{aligned}
f_{\mathbf e}(\mathbf x)
&=
\Pi_X\bigl(\Psi(\sigma_{\mathbf e}(\mathbf x))\bigr),\\
(\mathbf x,\mathbf e)
&\in
C_X\times C_E .
\end{aligned}
$$
Let
$$
\mathcal K_S
=
\{
\sigma_{\mathbf e}(\mathbf x):
\mathbf x\in C_X,\,
\mathbf e\in C_E
\}
\subset\mathcal Z .
$$
By the theorem, $\Psi$ can be approximated in $C^s$ on a compact
neighborhood of $\mathcal K_S$ by a \textsc{GRB-SympNet}
$\Phi_\theta^{\Delta t}$. Define the induced physical predictor
$$
\widehat f_{\theta,\mathbf e}
=
\Pi_X
\circ
\Phi_\theta^{\Delta t}
\circ
\sigma_{\mathbf e}.
$$
Then, for every $\varepsilon>0$, there exists a \textsc{GRB-SympNet} lifted
predictor such that
$$
\sup_{\mathbf e\in C_E}
\|
\widehat f_{\theta,\mathbf e}
-
f_{\mathbf e}
\|_{C^s(C_X)}
<
\varepsilon .
$$
At the same time, the lifted map remains exactly symplectic:
$$
\begin{aligned}
&D\Phi_\theta^{\Delta t}(\mathbf Z)^\top
\mathbf J_\Omega
D\Phi_\theta^{\Delta t}(\mathbf Z) = \mathbf J_\Omega .
\end{aligned}
$$
This formalizes the modeling principle used throughout \textsc{CaLiSym}. The
projected physical dynamics may be forced, dissipative, or contact-constrained,
while the learned map in the lifted canonical coordinates remains exactly
symplectic.

A sufficient local condition for the lifted-realization assumption is
that the joint map
$$
F(\mathbf x,\mathbf e)
=
(f_{\mathbf e}(\mathbf x),\mathbf e)
$$
is a local diffeomorphism. In the base--fiber coordinates used by the
lift,
$$
\begin{aligned}
\mathbf b
&=
(\mathbf q,\mathbf p,
\boldsymbol\mu_u,\boldsymbol\lambda_c),\\
\boldsymbol\zeta
&=
(\mathbf r,-\mathbf y,
-\boldsymbol\lambda_u,\boldsymbol\pi_c),\\
\Omega
&=
\sum_i db^i\wedge d\zeta_i,
\end{aligned}
$$
the cotangent lift $T^*F$ is symplectic and recovers the physical
transition after projection to the gauge-fixed data section. This is why
\textsc{CaLiSym} can impose symplecticity in $\mathcal Z$ without requiring the
projected map on $X$ to be symplectic.

\paragraph{KAN interpretation.}
The B-spline/KAN-style approximation is applied to the scalar ridge
Hamiltonians $K_\ell$, not componentwise to the vector-valued map
$\mathbf Z\mapsto\Phi_\theta^{\Delta t}(\mathbf Z)$. This distinction is
essential: componentwise approximation does not preserve
$$
D\Phi^\top\mathbf J_\Omega D\Phi
=
\mathbf J_\Omega .
$$
In \textsc{GRB-SympNet}, the learned spline functions enter only through scalar
Hamiltonians inside exact generalized-ridge flows, so expressivity is
added without breaking the symplectic constraint.

\vspace{-4mm}
\section{Training Details}
\label{app:training_details}
\subsection{Double pendulum training}
\label{app:pendulum}
\subsubsection{Hyperparameters}
Table~\ref{tab:training_hyperparameters} gives the configuration for the double pendulum, the low dimensional setting for the spline variant \textsc{GRB-SympNet}. The lifted phase space is $12$ dimensional and the model is small, $3{,}660$ parameters over $12$ layers. The scalar ridge Hamiltonian uses a cubic B-spline over $12$ intervals on $[-3.5, 3.5]$, evaluated in a projection dimension $m = 2$, which adds compact local expressivity in the ridge coordinate without breaking the exact symplectic flow. Training runs in FP64 with AdamW for $500$ epochs. The loss emphasizes the physical state ($w_q = 10$, $w_p = 5$) with a single control port ($w_\mu = 5$), since the pendulum is isolated with one torque port and no contacts.

\subsubsection{Data Preparation}

Unlike the quadruped and quadrotor benchmarks, which are derived from real-world datasets, the double-pendulum benchmark is generated directly from a known physical model. The system consists of a two-link planar pendulum subject to gravity, viscous damping, and external control torques. Trajectories are produced by numerically integrating the equations of motion using a high-order DOP853 integrator with zero-order-hold control inputs, yielding a controlled dissipative mechanical system with known ground-truth dynamics.

The state is represented using joint angles and generalized momenta. Since the configuration space is Euclidean, no manifold preprocessing is required. Control inputs are generated independently for each joint using randomized sinusoidal torque profiles, producing a diverse set of excitation patterns. Trajectories are generated directly at $100$ Hz and segmented into overlapping windows of $100$ time steps ($1$ s) with a stride of $10$ samples, resulting in approximately $40{,}000$ training windows and $1{,}600$ OOD windows.

To evaluate out-of-distribution generalization, the train and OOD splits are constructed using the system's total mechanical energy rather than a random partition of trajectories. OOD episodes are initialized in an energy regime approximately $36\%$ higher than that used for training, creating a physically meaningful distribution shift. Since mechanical energy directly governs the amplitude of motion, nonlinear coupling strength, and overall complexity of the resulting dynamics, higher-energy trajectories explore larger regions of phase space and exhibit qualitatively different long-horizon behavior than those observed during training.

Importantly, the distribution shift persists beyond the initial conditions. Although viscous friction continuously dissipates energy throughout each rollout, OOD trajectory windows exhibit approximately $56\%$ higher mean mechanical energy than training windows and consistently occupy more energetic regions of phase space throughout their evolution. Consequently, the benchmark evaluates generalization across distinct dynamical regimes and trajectory families rather than simple interpolation between nearby samples, yielding a substantially more challenging and physically grounded OOD evaluation protocol than conventional random trajectory splits.

\vspace{-4mm}
\subsection{Quadrotor training}
\label{app:quadrotor_hyperparameters}
\subsubsection{Hyperparameters} 
Table~\ref{tab:training_hyperparameters} gives the configuration for the quadrotor, the forced underactuated setting for \textsc{GR-SympNet} without contact ports. The lifted phase space is $36$ dimensional, realized with $23{,}688$ parameters. The lift embeds the rotor thrust and body torque commands as a single control port, so the contact block is omitted and energy enters only through actuation while drag acts as unmodeled dissipation. Training runs in FP64 with AdamW at a constant learning rate of $3\times10^{-3}$ for $500$ epochs, using a teacher forcing schedule held at $\alpha = 1.0$ for $50$ epochs followed by a $50$ epoch linear ramp. The loss emphasizes the physical state ($w_q = w_p = 10$) with light section and port weights ($w_{\mathrm{sec}} = w_\mu = 0.1$), in contrast to the heavily port weighted quadruped objective, since the quadrotor has a single smooth actuation channel rather than several heterogeneous ports.

\subsubsection{Data Preparation}
We evaluate the quadrotor experiments on the AscTec Pelican Flight Dataset~\cite{mohajerin18quad, moharejin19quadrnn}, which contains $54$ real-world flight trajectories collected on an AscTec Pelican platform. Unlike the quadruped, the quadrotor is a single free-flying rigid body without articulated joints or contacts, resulting in a substantially simpler preprocessing pipeline. The dataset provides pose, velocities, body rates, and motor commands at $100$ Hz. Each trajectory is expressed in local tangent-space coordinates by applying the logarithm map on $\mathbb{SE}(3)$ relative to the initial pose of each window, yielding a six-dimensional configuration vector $\boldsymbol{q}\in\mathbb{R}^6$.

Generalized momenta are computed as $\boldsymbol{p}=\mathbf{M}\boldsymbol{v}$ using the constant rigid-body mass matrix $\mathbf{M}=\mathrm{diag}(m\mathbb{I}_3,\mathbf{I}_{\mathrm{body}})$. Control inputs are represented as generalized wrenches obtained by mapping the four motor commands through the quadrotor allocation matrix, yielding $\boldsymbol{\tau}\in\mathbb{R}^6$.

The trajectories require no resampling or filtering and are segmented into overlapping windows of $100$ time steps ($1$ s) with a stride of $10$ samples. After a momentum-smoothness validation step removes anomalous segments, the resulting data are split into disjoint training and OOD flight sets to evaluate generalization to previously unseen trajectories.

\vspace{-4mm}
\subsection{Quadruped training}
\label{app:quadruped}
\subsubsection{Hyperparameters}
Table~\ref{tab:training_hyperparameters} gives the configuration for the quadruped, the high dimensional contact rich setting for \textsc{GR-SympNet}. The lifted phase space is $132$ dimensional, realized with $\sim\!316\mathrm{K}$ parameters over $18$ layers of width $66$. The generalized-ridge layers confine the nonlinear computation to a low dimensional ridge coordinate, which keeps the predictor parameter efficient despite the large phase space. Training runs in FP64 with Muon for $500$ epochs. The loss weights the control and contact ports heavily ($w_\mu = 65$, $w_{f_c} = 55$) against an equal state weighting ($w_q = w_p = 0.5$) and a light section penalty ($w_{\mathrm{sec}} = 0.01$). The heavy port weighting is the channel through which actuation work and contact impulses enter an otherwise conservative lifted flow.

\subsubsection{Data Preparation}
The \textsc{GrandTour} dataset~\cite{frey2026grandtour} comprises diverse real-world locomotion trajectories collected on the ANYmal-D quadruped~\cite{hutter16anymal} across a wide range of terrains, commands, gaits, and environmental conditions. The onboard state-estimation stack provides measurements at $400$ Hz. To construct a dataset compatible with the proposed lifted formulation, the raw estimator outputs are transformed into a canonical mechanical state representation and subsequently filtered, resampled, and validated.

The quadruped configuration consists of an $\mathbb{SE}(3)$ floating-base pose together with Euclidean joint coordinates. Since only the floating base evolves on a nonlinear manifold, each trajectory window is expressed relative to its initial base pose using the $\mathbb{SE}(3)$ logarithm map implemented in Pinocchio, while the joint coordinates remain unchanged. Generalized momenta are then computed from the rigid-body dynamics model yielding the canonical state representation $(\boldsymbol{q},\boldsymbol{p})$ used throughout this work.

The resulting trajectories are resampled from $400$ Hz to $100$ Hz. Prior to decimation, the momentum channels are filtered using an $18$ Hz sixth-order low-pass filter. Given the target Nyquist frequency of $50$ Hz, this removes high-frequency transients arising from impacts, contact transitions, and estimator noise while preserving the dominant locomotion dynamics, which typically lie below $10$ Hz. This prevents aliasing artifacts from contaminating the momentum channels after downsampling.

Finally trajectories are segmented into overlapping windows of $100$ time steps ($1$ s) with a stride of $10$ samples. A validation stage rejects windows exhibiting anomalous momentum discontinuities, and the remaining data are partitioned into disjoint training and OOD mission sets to evaluate generalization across previously unseen operating conditions.
\vspace{-3mm}
\begin{table}[ht]
\centering
\caption{Training configurations for the double pendulum, quadrotor, and quadruped experiments.}
\label{tab:training_hyperparameters}
\renewcommand{\arraystretch}{1.05}
\begin{tabular}{lccc}
\toprule
 & \textbf{Double Pend.} & \textbf{Quadrotor} & \textbf{Quadruped} \\
\midrule
\multicolumn{4}{l}{\textbf{Architecture}} \\
\midrule
\textsc{SympNet} type & GRB & GR & GR \\
Phase-space dim. & 12 & 36 & 132 \\
Parameters & 3{,}660 & 23{,}688 & 316{,}008 \\
Layers ($L$) & 12 & 14 & 18 \\
Width ($W$) & -- & 36 & 66 \\
Activation & -- & $\tanh$ & $\tanh$ \\
Projection dimension & 2 & -- & -- \\
Spline degree & 3 & -- & -- \\
Spline intervals & 12 & -- & -- \\
Knot range & $[-3.5,\,3.5]$ & -- & -- \\
Backbone hidden dim. & 8 & - & - \\
\midrule
\multicolumn{4}{l}{\textbf{Dataset}} \\
\midrule
Training windows & 40{,}000 & 98{,}552 & 126{,}309 \\
ID test windows & 1{,}000 & 24{,}638 & 31{,}577 \\
OOD test windows & 1{,}600 & 15{,}129 & 14{,}780 \\
Sampling period ($\delta t$) [s] & 0.01 & 0.01 & 0.01 \\
Evaluation horizon ($\Delta \mathrm T$) & 200 ($2s$) & 100 ($1s$) & 100 ($1s$) \\
\midrule
\multicolumn{4}{l}{\textbf{Training}} \\
\midrule
Optimizer & AdamW & AdamW & Muon \\
Learning rate & $5\times10^{-3}$ & $3\times10^{-3}$ & $10^{-2}$ \\
Weight decay & $2\times10^{-4}$ & $10^{-3}$ & $10^{-3}$ \\
Gradient clipping & 100 & 100 & 500 \\
Precision & FP64 & FP64 & FP64 \\
Epochs & 500 & 500 & 500 \\
Batch size & 1024 & 1024 & 2048 \\
TF training horizon & 100 & 100 & 100 \\
\midrule
\multicolumn{4}{l}{\textbf{Loss Weights}} \\
\midrule
Configuration ($w_q$) & 10.0 & 10.0 & 0.5 \\
Momentum ($w_p$) & 5.0 & 10.0 & 0.5 \\
Section ($w_{\mathrm{sec}}$) & 5.0 & 0.01 & 0.01 \\
Control port ($w_\mu$) & 5.0 & 0.1 & 65.0 \\
Contact port ($w_{f_c}$) & -- & -- & 55.0 \\
\midrule
Symplectic normalization & Enabled & Enabled & Enabled \\
\bottomrule
\end{tabular}
\end{table}

\subsection{Training Hardware}
All models were trained in double precision using PyTorch inside an NVIDIA NGC container. Full training runs and the hyperparameter searches that selected each configuration executed on NVIDIA GH200 Grace Hopper superchips. Early development and debugging ran locally on a single NVIDIA RTX~4090.

\vspace{-2mm}
\bibliography{references}

\end{document}